\pdfoutput=1
\pdfoutput=1
\documentclass[11pt]{article}

\usepackage[margin=1in]{geometry}
\usepackage[T1]{fontenc}
\usepackage[utf8]{inputenc}
\usepackage{newtxtext}
\usepackage{newtxmath}
\usepackage{microtype}
\usepackage{authblk}
\usepackage[hidelinks,hypertexnames=false]{hyperref}
\usepackage[table]{xcolor}
\hypersetup{colorlinks=true, linkcolor=blue, citecolor=blue, urlcolor=blue}
\usepackage{graphicx}
\setkeys{Gin}{draft=false}
\usepackage{float}
\usepackage{wrapfig}
\usepackage{array}
\usepackage{booktabs}
\usepackage{ragged2e}
\usepackage{siunitx}
\usepackage{makecell}
\usepackage{pgfplots}
\pgfplotsset{compat=1.18}
\usepackage{caption}
\usepackage[square,numbers,sort]{natbib}

\makeatletter
\newcommand{\appendixtoc}{%
  \begingroup%
  \setcounter{tocdepth}{2}%
  \section*{Table of Contents}%
  \@starttoc{app}%
  \endgroup%
}
\newcommand{\appsection}[1]{%
  \section{#1}%
  \addcontentsline{app}{section}{\protect\numberline{\thesection}#1}%
}
\newcommand{\appsubsection}[1]{%
  \subsection{#1}%
  \addcontentsline{app}{subsection}{\protect\numberline{\thesubsection}#1}%
}
\makeatother

\newcommand{\takeawaybox}[1]{%
  \vspace{0.4\baselineskip}%
  \noindent\fbox{\parbox{0.97\linewidth}{#1}}%
  \vspace{0.2\baselineskip}%
}

\title{\textbf{FLUX: Data Worth Training On}}
\author{\textbf{Gowtham} \quad \textbf{Sai Rupesh} \quad \textbf{Sanjay Kumar} \quad \textbf{Saravanan} \quad \textbf{Venkata Chaithanya}}
\date{}

\begin{document}
\begin{center}
\rule{\linewidth}{2.2pt}

{\LARGE\bfseries FLUX: Data Worth Training On\par}
\vspace{0.6em}
{\large \textbf{Gowtham} \quad \textbf{Sai Rupesh} \quad \textbf{Sanjay Kumar} \quad \textbf{Saravanan} \quad \textbf{Venkata Chaithanya}\par}
\vspace{0.4em}
{\small \texttt{contact@blubridge.ai}\par}

\rule{\linewidth}{0.6pt}
\end{center}

\begin{abstract}
\justifying\noindent
Modern large language model training is no longer limited by data availability, but by the inability of existing preprocessing pipelines to simultaneously achieve massive scale and high data quality. Current approaches are forced to sacrifice one for the other---either aggressively filtering to improve quality at the cost of severe token loss, or retaining large volumes of data while introducing substantial noise. In this work, we introduce \textbf{FLUX}, a preprocessing pipeline specifically designed to break this long-standing trade-off by maximizing token retention while enforcing rigorous quality control. Models trained on FLUX-curated data consistently outperform prior methods. A \textbf{3B-parameter} model trained on \textbf{60B tokens} with FLUX achieves \textbf{32.14\%} MMLU accuracy, surpassing the previous state-of-the-art pipeline DCLM (\textbf{31.98\%}) and significantly outperforming FineWeb (\textbf{29.88\%}). FLUX achieves the same aggregate score as a model trained on DCLM data using only \textbf{39B tokens}, resulting in a \textbf{34.4\%} reduction in training compute. At the data level, FLUX extracts \textbf{50B} usable tokens from a single dump (CC-MAIN-2025-51), compared to \textbf{40B} from DCLM (\textbf{+25\%} retention). \textbf{FLUX-Base} yields \textbf{192B} tokens, exceeding FineWeb's \textbf{170B} while still maintaining superior quality. Overall, FLUX establishes a new state of the art in web-scale data preprocessing by demonstrating that high retention, strong quality control, and computational efficiency can be achieved simultaneously, redefining the limits of scalable dataset construction for modern language models.
\end{abstract}

\section{Introduction}
While much of the research on LLMs has focused on scaling model parameters and computational resources, the quality, composition, and diversity of training data play a crucial role in determining model performance and generalization~\citep{raffel_exploring_2023,hoffmann_training_2022,kaplan_scaling_2020}. However, web-scale corpora are often saturated with noise, duplication, and low-information content~\citep{lee_deduplicating_2022,soldaini_dolma_2024}, which degrade training efficiency, destabilize optimization~\citep{lee_deduplicating_2022}, and ultimately limit downstream performance. The central challenge, therefore, is not merely raw data availability, but the design of preprocessing pipelines~\citep{penedo_fineweb_2024,li_datacomp-lm_2025} that preserve massive scale while enforcing strict quality control.

Existing data curation pipelines such as \textbf{DataComp for Language Models (DCLM)~\citep{li_datacomp-lm_2025}} and our previous work \textbf{Blu-WERP}~\citep{gowtham_blu-werp_2025} achieve \textbf{strong downstream performance by producing high-quality datasets}. However, these improvements often come at the cost of \textbf{reduced token retention}~\citep{penedo_fineweb_2024,li_datacomp-lm_2025}. For instance, DCLM retains only 3.8 trillion tokens from 89 Common Crawl dumps, indicating \textbf{substantial data loss during preprocessing}~\citep{li_datacomp-lm_2025}. Importantly, this does not represent a saturation point; there remains significant scope to further improve both \textbf{token retention and dataset quality} through more effective preprocessing strategies.

To address the challenges present in existing pipelines~\citep{li_datacomp-lm_2025,penedo_fineweb_2024}, we introduce FLUX, a preprocessing pipeline for large-scale language model training data curation. FLUX is designed to jointly optimize data quality and token retention, enabling the construction of large, high-quality datasets suitable for modern LLM training. Importantly, the pipeline emphasizes \textbf{computational efficiency}, ensuring that improvements in dataset quality and scale do not introduce prohibitive compute overhead.

\begin{wrapfigure}{r}{0.42\textwidth}
  \vspace{-0.4\baselineskip}
  \centering
  \includegraphics[width=0.40\textwidth]{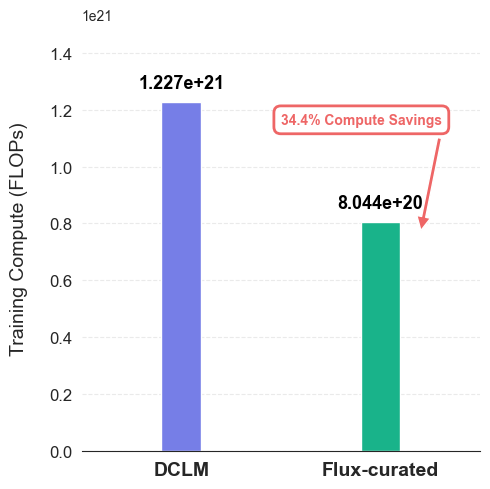}
  \caption{\textbf{Compute Savings at Equal Performance (3B scale).} DCLM requires $1.227\times10^{21}$ FLOPs to reach 50.48\% aggregate score, whereas FLUX reaches the same performance with $8.044\times10^{20}$ FLOPs (about 34.4\% lower compute).}
  \label{fig:compute_efficiency_chinchilla}
  \vspace{-0.8\baselineskip}
\end{wrapfigure}

We conduct experiments across model sizes ranging from \textbf{530M to 3B parameters} using the OLMo training framework~\citep{groeneveld_olmo_2024}, following the Chinchilla scaling rule with a 20$\times$ token-to-parameter ratio~\citep{hoffmann_training_2022}. For the \textbf{3B model}, training on the DCLM dataset requires 1.227 $\times$ 10$^{21}$ FLOPs to achieve an aggregate score of 50.48\%. In contrast, the model trained on the \textbf{FLUX-curated dataset reaches the same performance level using only 8.044 $\times$ 10$^{20}$ FLOPs}, resulting in approximately \textbf{34.4\% compute savings} compared to training on DCLM. Notably, this \textbf{trend is consistent across model scales}, with similar improvements observed for models ranging from 530M to 3B parameters.

One of the primary objectives of \textbf{FLUX} is to curate \textbf{high-quality datasets while minimizing computational overhead}. For parsing, FLUX uses \textbf{Apex}, a lightweight and efficient alternative to heavier extractors, enabling \textbf{high-quality text extraction with substantially lower runtime overhead}. This parser-level efficiency is summarized in Table~\ref{tab:parser-comparison}.

Beyond parser choice, FLUX emphasizes a \textbf{minimal yet effective filtering} strategy that preserves token retention while improving dataset quality. Combined with a Rust-based implementation optimized for large-scale processing, this design yields \textbf{strong end-to-end preprocessing efficiency across competing pipelines}, as shown in Table~\ref{tab:intro_pipeline_efficiency}.

\begin{table}[H]
  \centering
  \small
  \setlength{\tabcolsep}{8pt}
  \renewcommand{\arraystretch}{1.2}
  \begin{tabular}{l c c c c c}
    \toprule
    \textbf{Pipeline} & \textbf{Input} & \textbf{Wall-clock} & \textbf{CPU-hrs} & \textbf{CPU-hrs/TB} & \textbf{Vs. FLUX} \\
    \midrule
    DCLM & 1.20 TB & 11 hr 7 min & 355.7 & 296.4 & 6.26x \\
    FineWeb & 0.60 TB & 5 hr 39 min & 181 & 301.3 & 6.36x \\
    \rowcolor{blue!10} FLUX & 1.03 TB & 1 hr 31 min & 48.5 & 47.3 & 1.00x \\
    \bottomrule
  \end{tabular}
  \caption{\textbf{Pipeline-level preprocessing efficiency comparison.}}
  \label{tab:intro_pipeline_efficiency}
\end{table}

\section{Related Work}

Existing preprocessing pipelines for Common Crawl differ in filtering, deduplication, and classification strategies, but their core design is largely homogeneous. Most rely on generalized heuristic rules---such as repetition filtering, language identification, and coarse quality thresholds---to remove low-information content. Differences across pipelines primarily reflect variations in filtering aggressiveness rather than fundamentally distinct architectural approaches~\citep{li_datacomp-lm_2025,penedo_fineweb_2024,gohari_gneissweb_2025,gowtham_blu-werp_2025,soldaini_dolma_2024}.

Most large-scale data curation pipelines begin with an \textbf{HTML parsing stage} rather than relying directly on WET files, as parsing provides greater control over text extraction and allows for quality improvements early in the preprocessing process~\citep{li_datacomp-lm_2025,penedo_fineweb_2024,gowtham_blu-werp_2025}. Parser choice introduces a trade-off between token retention, extraction quality, and computational cost. For example, Trafilatura prioritizes higher output quality, often at the expense of higher computational overhead and lower retention, whereas Resiliparse emphasizes computational efficiency and higher token retention but compromises on quality~\citep{barbaresi_trafilatura_2021,bevendorff_elastic_chatnoir_2021}; this comparison is summarized in Table~\ref{tab:parser-comparison}.

\textbf{Undesirable data} in large-scale web corpora can broadly be classified into two categories: \textbf{noise} and \textbf{semantically weak content}. Many data curation pipelines employ \textbf{heuristic filters} to remove noisy content from web-scale datasets~\citep{li_datacomp-lm_2025,penedo_fineweb_2024,gohari_gneissweb_2025}. The key question, however, lies in \textbf{how these filters are applied and configured} within the pipeline. The main challenge with heuristic filters is finding the right trade-off between \textbf{token retention} and \textbf{dataset quality}~\citep{li_datacomp-lm_2025,penedo_fineweb_2024,gowtham_blu-werp_2025}. Blu-WERP applies aggressive heuristic filtering, and although it uses Justext as the parser instead of Resiliparse, the advantage at the parser stage is largely nullified by the \textbf{aggressive filtering}~\citep{gowtham_blu-werp_2025}. As a result, the dataset quality achieved after heuristic filtering is close to that of DCLM. This occurs because Blu-WERP relies heavily on \textbf{document-level filters}. As a result of removing it, more tokens are removed, which reduces data diversity and ultimately affects the aggregate score. The key factor is not merely how filtering is performed, but \textbf{which filters and their respective thresholds are used}~\citep{gowtham_blu-werp_2025,li_datacomp-lm_2025}.

\begin{figure}[H]
\centering
\begin{minipage}{0.49\textwidth}
    \centering
    \includegraphics[width=\linewidth]{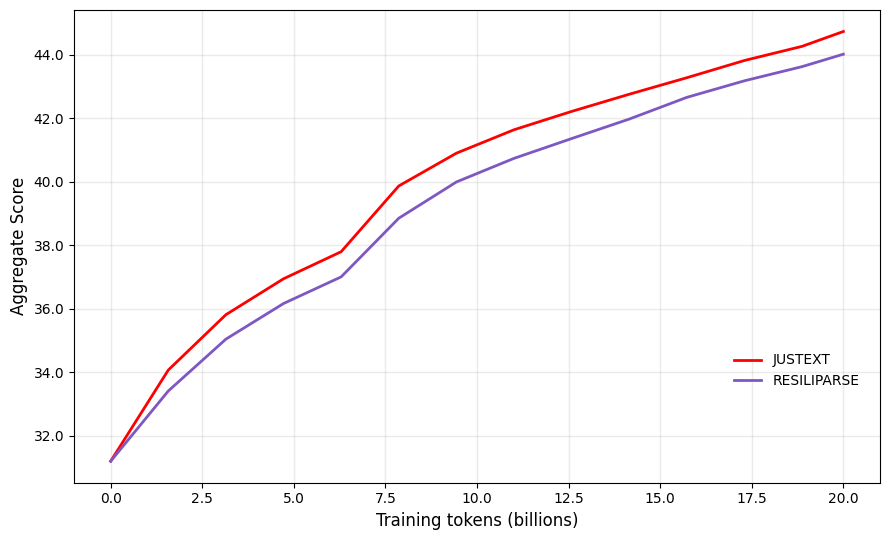}
    \caption{\textbf{Aggregate performance after the parsing stage.} Blu-WERP exhibits a measurable advantage over DCLM immediately following text extraction.}
    \label{fig:related_work_1}
\end{minipage}\hfill
\begin{minipage}{0.49\textwidth}
    \centering
    \includegraphics[width=\linewidth]{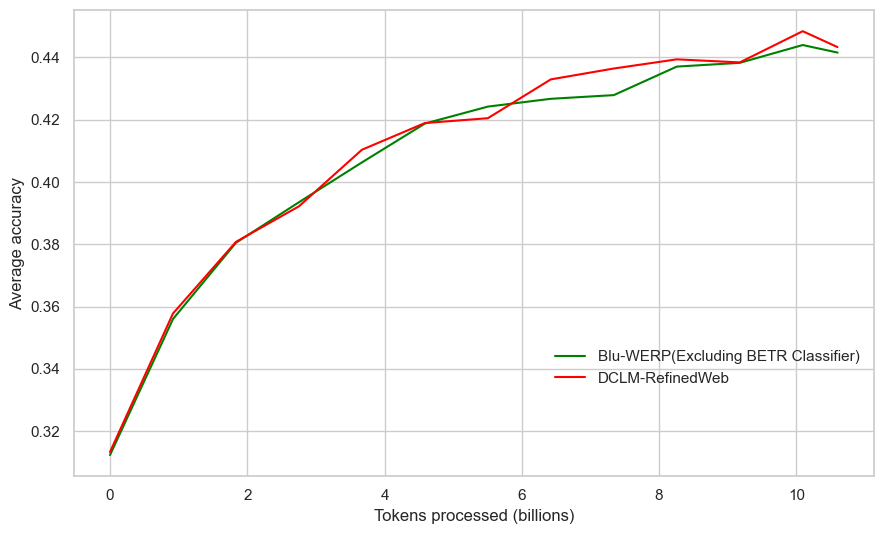}
    \caption{\textbf{Aggregate performance after heuristic filtering.} The aggressive filtering stage substantially attenuates the parsing-stage advantage observed for Blu-WERP.}
    \label{fig:related_work_2}
\end{minipage}
\end{figure}

Semantically weak content cannot be effectively removed from web corpora using \textbf{heuristic filters alone}; it typically requires \textbf{model-based classification}~\citep{li_datacomp-lm_2025,penedo_fineweb_2024,gohari_gneissweb_2025}. There are multiple ways to apply model-based filtering, depending on the priorities of a pipeline in terms of \textbf{compute cost}, \textbf{token retention}, and dataset quality. One approach uses LLMs as classifier after annotating datasets, as seen in \textbf{FineWeb-Edu}~\citep{penedo_fineweb_2024}. However, this approach results in very \textbf{low retention} (often below 10\%) and incurs high computational cost. On the other hand, lighter models based on n-gram features, such as \textbf{FastText}, can classify the same data in less than an hour, making them significantly more computationally efficient~\citep{joulin_bag_2016,gohari_gneissweb_2025}. Within FastText-based approaches, additional strategies can be used to reduce false positive and false negative rates, such as the \textbf{Ensemble Quality filtering method employed by GneissWeb}~\citep{gohari_gneissweb_2025}.

\begin{table}[H]
  \centering
  \small
  \setlength{\tabcolsep}{10pt}
  \renewcommand{\arraystretch}{1.15}
  \begin{tabular}{l c}
    \toprule
    \textbf{Component} & \textbf{Classification time (in mins)} \\
    \midrule
    FastText (Single-bin) & 30 \\
    FastText (Dual-bin) & 60 \\
    BERT (110M) & 1290 \\
    \bottomrule
  \end{tabular}
  \caption{\textbf{Time taken by the components to classify 100B tokens from CC-MAIN-2025-51 snapshot post deduplication.} The time taken by the FastText also depends on the size of the bin. We used bin sizes of 1.7GB for both single and dual-bin as mention in Section \ref{sec:dual_bin_quality}.}
  \label{tab:related_work_classification_time}
\end{table}

\section{Building High-Quality Training Data with FLUX}

\subsection{Text Extraction}

Common Crawl provides web data in three primary formats: WARC, WET, and WAT~\citep{soldaini_dolma_2024}. WARC files preserve raw HTTP responses and HTML content, WET files contain pre-extracted plain text, and WAT files store structured metadata. For large-scale pretraining, we process WARC data directly to retain full control over extraction behavior, filtering criteria, and final text quality.

In our extraction pipeline, we benchmark Apex, Resiliparse~\citep{bevendorff_elastic_chatnoir_2021}, and Trafilatura~\citep{barbaresi_trafilatura_2021} on 10K WARC files using a \texttt{c8a.8xlarge} instance and the \texttt{o200k\_base} tokenizer. In addition, we apply targeted post-extraction filtering to improve textual cleanliness and consistency in the final corpus.

As shown in Table~\ref{tab:parser-comparison}, Apex and Resiliparse~\citep{bevendorff_elastic_chatnoir_2021} deliver nearly identical token yield, while Apex provides lower runtime. Trafilatura~\citep{barbaresi_trafilatura_2021} is substantially slower and yields fewer tokens. Given this efficiency--coverage trade-off, we adopt Apex as the primary parser in the production pipeline, with additional downstream filtering to maintain corpus quality.

\begin{table}[htbp]
\centering
\small
\setlength{\tabcolsep}{6pt}
\renewcommand{\arraystretch}{1.12}
\begin{tabular}{>{\raggedright\arraybackslash}p{2.2cm}cccc}
\toprule
\textbf{Parser} & \makecell{\textbf{Total Tokens}\\\textbf{(Multilingual)}} & \makecell{\textbf{Total Tokens}\\\textbf{(English)}} & \textbf{Aggregate} & \makecell{\textbf{Compute Cost}\\\textbf{(hrs)}} \\
\midrule
Resiliparse & 279.4B & \textbf{54.6B} & 0.4089 & 6.63 \\
\rowcolor{blue!10} Apex & \textbf{279.5B} & 54.0B & 0.4123 & \textbf{4.31} \\
Trafilatura & 136.7B & 35.5B & \textbf{0.4330} & 47.33 \\
\bottomrule
\end{tabular}
\caption{\textbf{Parser Comparison with Runtime Cost (10K WARC files from CC-MAIN-2025-51 snapshot; 530M scale; 10.6B tokens).} Total multilingual/English token yield, aggregate downstream score, and compute cost (hours) across Resiliparse, Apex, and Trafilatura.}
\label{tab:parser-comparison}
\end{table}

\begin{figure}[H]
\centering
\includegraphics[width=0.90\linewidth]{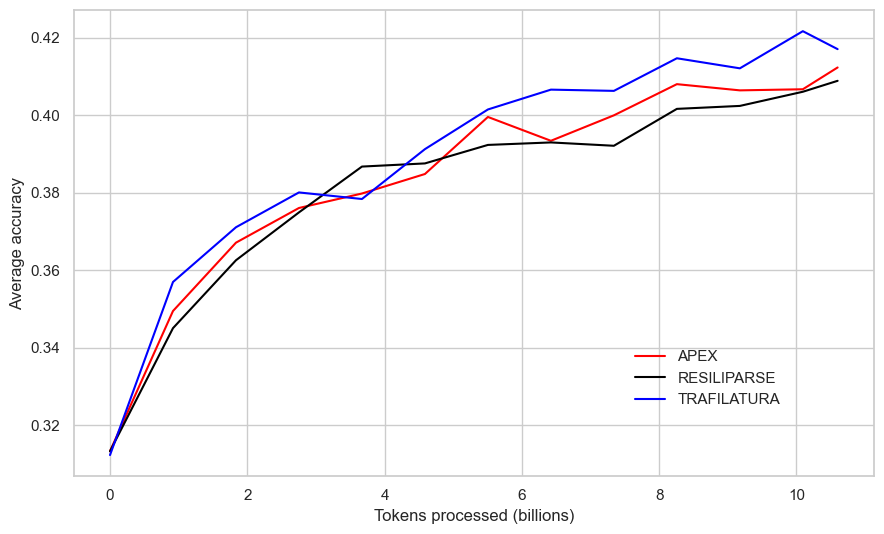}
\caption{\textbf{Parser-Level Downstream Performance Comparison (530M scale; 10.6B tokens).} This figure reports aggregated downstream benchmark performance for models trained on corpora produced by each parser configuration (Apex, Resiliparse, and Trafilatura) under the same training setup. It highlights the quality--retention trade-off across parsers and motivates the final parser choice used in the FLUX pipeline.}
\label{fig:parser-aggregated-result}
\vspace{-0.3em}
\end{figure}
\vspace{-0.2em}
\subsection{Heuristic Filtering}
\label{sec:heuristic_filtering}

FLUX applies a deterministic, multi-stage filtering pipeline governed by a single design principle: prefer intra-document line-level excision over whole-document rejection wherever feasible. Prior pipelines such as DCLM [3] enforce binary document-level decisions~\citep{penedo_refinedweb_2023,li_datacomp-lm_2025} at most filtering stages, which improves filter precision but systematically discards high-quality content embedded in otherwise noisy documents. FLUX instead excises individual low-quality lines and retains the cleaned body, recovering signal that binary rejection would permanently lose.

The filtering pipeline is partitioned into four sequential functional groups: (i) URL-based pre-filtering, which removes documents from domains associated with adult content prior to any content analysis~\citep{penedo_refinedweb_2023}; (ii) language identification, which retains English documents at a whole-document confidence threshold of >= 0.65 using FastText \texttt{lid.176.bin} ~\citep{joulin_bag_2016,joulin_fasttextzip_2017}, with non-English content routed to a multilingual partition rather than discarded; (iii) document-level quality gating via Gopher Quality~\citep{gopher2021}, Gopher Repetition, Nemo, and a Custom Quality Filter (stop-word ratio, unclosed-bracket ratio, minimum token count); and (iv) line-level cleaning via a content transformer that excises boilerplate lines across eleven heuristic classes, followed by a Word Removal Ratio filter that rejects documents from which more than 5\% of tokens were removed, guarding against documents too contaminated to yield coherent training signal. Full stage-by-stage specifications and thresholds are provided in Appendix~\ref{app:filter_pipeline}.

\begin{table}[H]
  \centering
  \small
  \begin{tabular}{lcc}
    \toprule
    \textbf{Pipeline} & \textbf{Post-Dedup Tokens} & \textbf{Aggregate (530M)} \\
    \midrule
    FineWeb & 24.50B & 44.31 \\
    DCLM & 24.81B & 44.40 \\
    \rowcolor{blue!10} FLUX & 27.22B & 45.45 \\
    \bottomrule
  \end{tabular}
  \caption{\textbf{FLUX vs DCLM-RefinedWeb and FineWeb --- filtering stage results (530M scale; 10.6B tokens).}}
  \label{tab:heuristic_filtering_flux_vs_baselines}
\end{table}

\textbf{Pipeline ablation.} To validate this design, we construct four configurations of increasing sophistication --- URL+LID, DocFilter, LineClean, and FLUX --- and evaluate both post-deduplication token yield and downstream aggregate score at the 530M scale (Appendix~\ref{app:filter_pipeline}, Section~\ref{app:filter_d3}). URL+LID applies only URL filtering and language identification, yielding 42.4B tokens post-deduplication with an aggregate score of 41.76. DocFilter adds the standard document-level filter suite (Gopher, Nemo, C4 Badwords~\citep{raffel_exploring_2023}); tokens collapse to 23.4B (-44.8\%) while aggregate improves to 44.24, confirming that quality filtering matters but at severe retention cost. LineClean introduces Custom Quality Filter and targeted line-level cleaning, achieving 44.68 aggregate at 21.0B tokens --- improved quality but further reduced retention from stacking document-level and line-level rejection simultaneously. FLUX removes C4 Badwords~\citep{raffel_exploring_2023} entirely and enforces those noise classes exclusively at line resolution with six additional targeted heuristics, recovering 27.2B post-deduplication tokens --- 16.4\% more than DocFilter --- while achieving the highest aggregate score of 45.45~\citep{li_datacomp-lm_2025,penedo_fineweb_2024}.

As reported in Table~\ref{tab:heuristic_filtering_flux_vs_baselines}, FLUX is the only configuration to simultaneously exceed both DCLM and FineWeb on token retention and downstream quality. FLUX retains 9.7\% more tokens than DCLM and 11.1\% more than FineWeb after deduplication, while outperforming both on aggregate score at the 530M scale. This demonstrates that the retention--quality trade-off is not a fundamental constraint of web-scale filtering~\citep{li_datacomp-lm_2025,penedo_fineweb_2024}. With the right design, both dimensions improve together.

\takeawaybox{\textbf{Baseline clarification.} We evaluate \textbf{FLUX} AND \textbf{FLUX-Base} against two reference configurations derived from the DCLM pipeline. The first, which we refer to as \textbf{DCLM}, corresponds to the full published pipeline of \citet{li_datacomp-lm_2025}: HTML extraction with Resiliparse~\citep{noauthor_chatnoir-euchatnoir-resiliparse_2026}, RefinedWeb-style heuristic filtering~\citep{penedo_refinedweb_2023}, Bloom filter deduplication ~\citep{noauthor_allenaibff_2025}, and a FastText classifier trained on OpenHermes 2.5~\citep{teknium_openhermes_2_5} and ELI5~\citep{fan_eli5_2019} data. The second, which we refer to as \textbf{DCLM-RefinedWeb}, omits the classification stage entirely, retaining only the extraction, filtering, and deduplication components. This separation allows us to attribute performance differences specifically to the classifier, rather than to the upstream pipeline as a whole.}

\begin{figure}[H]
  \centering
  \includegraphics[width=0.92\linewidth]{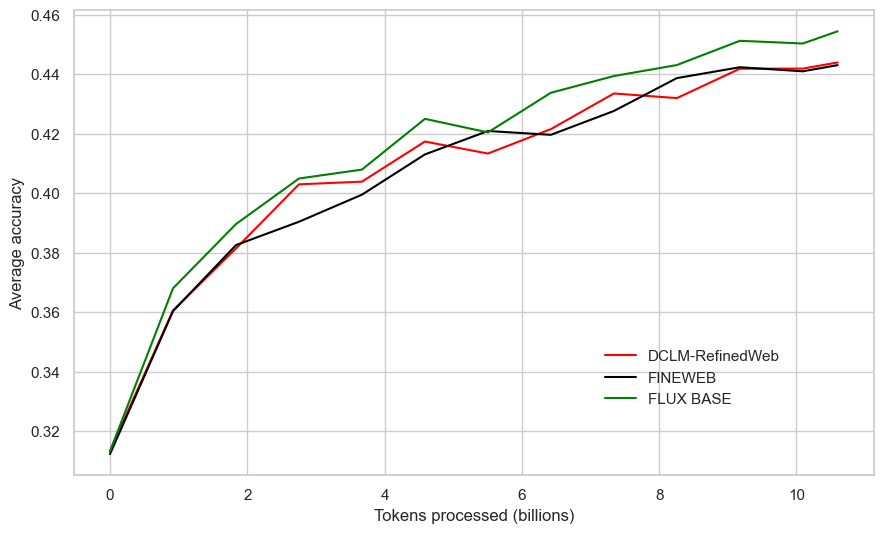}
  \caption{\textbf{Heuristic filtering summary (530M scale; 10.6B tokens).} Overview figure for the FLUX filtering design and its retention--quality behavior.}
  \label{fig:sec32_main_figure}
\end{figure}

\subsection{Deduplication}
\label{sec:deduplication}
We apply Bloom Filter-based Fuzzy Deduplication (BFF)~\citep{bloom_spacetime_1970} at the paragraph and document level to remove near-duplicate content~\citep{lee_deduplicating_2022} across dumps. Paragraphs are shingled into fixed-length n-grams~\citep{broder1997resemblance} and compared against a shared Bloom filter; highly duplicate paragraphs are dropped, and documents with sufficient duplicate paragraphs are removed in the same pass. Deduplication is run globally across both dumps with a single shared filter to capture cross-dump redundancy. Full pipeline details, parameters, and corpus-scale statistics are provided in Appendix~\ref{app:dedup}.

\subsection{Model-Based Classification}
\label{sec:dual_bin_quality}

\paragraph{Dual-Bin FastText Classification}
FastText supervised classifiers have been widely adopted~\citep{joulin_bag_2016} for document-level quality filtering in web-scale pretraining pipelines~\citep{li_datacomp-lm_2025,penedo_fineweb_2024}, owing to their sub-linear inference cost and demonstrated effectiveness against significantly heavier model-based alternatives. Under a single-classifier design, quality and retention are coupled through a single decision threshold: raising the threshold improves precision at the direct cost of recall, forcing a zero-sum trade-off. A dual-bin framework breaks this rigidity~\citep{gohari_gneissweb_2025}---by deploying two independently calibrated classifiers under a logical-OR acceptance rule, documents rejected by one bin may be recovered by the other, achieving strictly superior quality-retention operating points relative to any single-bin policy at equivalent thresholds.

We deploy two classifiers under a logical-OR framework. The first is the off-the-shelf FastText classifier introduced by the DCLM pipeline~\citep{li_datacomp-lm_2025}, trained on OpenHermes 2.5~\cite{teknium_openhermes_2_5} and ELI5~\cite{fan_eli5_2019} data. We denote this base classifier as $\phi_{\mathrm{DCLM}}$. The second is a FastText classifier, $\phi_{\mathrm{BETR}}$, constructed following benchmark-targeted ranking (BETR)~\citep{mizrahi_language_2025}, which operationalizes document quality as embedding-space proximity to downstream benchmark examples.

\paragraph{BETR bin ($\phi_{\mathrm{BETR}}$).}
BETR~\citep{mizrahi_language_2025} demonstrated that scoring pretraining documents by their maximum cosine similarity to benchmark train examples---and distilling these scores into a lightweight FastText classifier---yields consistent downstream gains while remaining computationally tractable at web scale. Critically, BETR established that a max-similarity scoring function, which assigns each document a score equal to its highest per-example similarity across all benchmarks, strictly outperforms mean-similarity aggregation, and that a FastText classifier trained on this signal matches or exceeds substantially larger LM-based classifiers on downstream evaluation.

Following this methodology, we embed the train splits of all nine evaluation benchmarks---MMLU \cite{hendrycks_measuring_2021}, HellaSwag \cite{noauthor_hellaswag_nodate}, ARC-Easy, ARC-Challenge \cite{clark_think_2018}, CSQA \cite{talmor_commonsenseqa_2019}, PIQA \cite{bisk_piqa_2019}, SocialIQA \cite{sap_socialiqa_2019}, WinoGrande \cite{sakaguchi_winogrande_2019}, and OpenBookQA \cite{mihaylov_can_2018}---alongside approximately 1M documents sampled uniformly from our post-deduplication corpus, using Arctic-Embed L v2~\citep{merrick_arctic-embed_2024} as the shared embedding model. Each document is assigned a representative score equal to its maximum cosine similarity across all benchmark examples and all benchmarks. A balanced training set of 200K documents is constructed as follows:

\begin{itemize}
  \item \textbf{Positive documents (\textasciitilde100K):} Documents in the top 10\% by maximum cosine similarity score---content semantically proximate to the linguistic and reasoning patterns present in benchmark train examples.
  \item \textbf{Negative documents (\textasciitilde100K):} Documents uniformly sampled from the remaining 90\% of the scored corpus, following the negative construction protocol of BETR~\citep{mizrahi_language_2025}.
\end{itemize}

A supervised FastText classifier~\citep{joulin_bag_2016} is trained on this 200K set with the following hyperparameters: learning rate 0.1, embedding dimension 100, context window size 5, bigram features (wordNgrams = 2), softmax loss, minimum word count threshold 1, and 5 training epochs. A fixed random seed of 42 is used throughout for reproducibility.

\paragraph{Acceptance rule.}
Given a document $d$, let $s_{\mathrm{DCLM}}(d)$ and $s_{\mathrm{BETR}}(d)$ denote the scalar confidence scores produced by $\phi_{\mathrm{DCLM}}$ and $\phi_{\mathrm{BETR}}$ respectively, each calibrated to $[0, 1]$ for the positive (high-quality) class label. Document $d$ is accepted into the final corpus if and only if:
$$
s_{\mathrm{DCLM}}(d) \ge \tau_{\mathrm{DCLM}} \;\; \lor \;\; s_{\mathrm{BETR}}(d) \ge \tau_{\mathrm{BETR}}
$$

\paragraph{Threshold selection.}
Operating thresholds are selected via a systematic sweep over $(\tau_{\mathrm{DCLM}}, \tau_{\mathrm{BETR}})$ pairs evaluated at the 530M parameter scale. Notably, while the original DCLM pipeline operates $\phi_{\mathrm{DCLM}}$ at a standalone threshold of 0.018112, deploying it in a dual-bin logical-OR architecture requires recalibration. As shown in our sweep (Table~\ref{tab:threshold_sweep}), pairing the original DCLM threshold ($\tau_{\mathrm{DCLM}} = 0.018112$) \citep{li_datacomp-lm_2025} with BETR yields high retention but suboptimal downstream accuracy. 

\begin{figure}[H]
\centering
\includegraphics[width=0.95\linewidth]{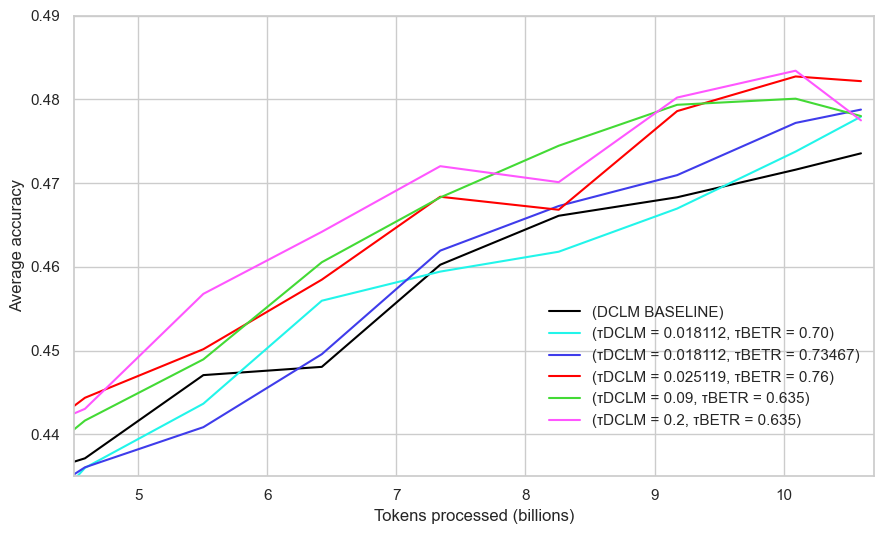}
\caption{\textbf{Dual-Bin Threshold Sweep and Final Operating Point (530M scale; 10.6B tokens).} Aggregate learning curves across all evaluated $(\tau_{\mathrm{DCLM}},\tau_{\mathrm{BETR}})$ settings on a fixed 100B post-deduplication input from CC-MAIN-2025-51. The selected operating point $(\tau_{\mathrm{DCLM}}=0.025119,\tau_{\mathrm{BETR}}=0.76)$ consistently leads after early training and yields the best final Aggregate while retaining 26.88\% of input tokens.}
\label{fig:classifier_curves}
\end{figure}

Instead, the configuration $(\tau_{\mathrm{DCLM}} = 0.025119, \tau_{\mathrm{BETR}} = 0.76)$ is selected. By tightening the DCLM threshold to 0.025119, we filter out slightly more marginal general-knowledge documents, making room to recover high-value reasoning documents via the BETR bin. This operating point produces the highest Aggregate score and MMLU among all evaluated pairs, as evidenced by the learning curves in Figure~\ref{fig:classifier_curves}.

The token-level impact of the dual-bin classifier on CC-MAIN-2025-51 is summarized in Table~\ref{tab:dual_bin_retention_cc51}. FLUX extracts 50B usable tokens from this dump compared to 40B under the DCLM pipeline---a 25\% improvement in token retention---while simultaneously achieving higher downstream model quality.

\begin{table}[H]
  \centering
  \small
  \setlength{\tabcolsep}{10pt}
  \renewcommand{\arraystretch}{1.15}
  \begin{tabular}{l c}
    \toprule
    \textbf{Pipeline} & \textbf{Tokens Retained} \\
    \midrule
    DCLM & 40B \\
    \rowcolor{blue!10} FLUX & 50B \\
    \bottomrule
  \end{tabular}
  \caption{\textbf{Token retention after Dual-Bin FastText classification (CC-MAIN-2025-51).} FLUX retains 50B tokens from this dump compared to 40B for DCLM---a 25\% improvement---while simultaneously achieving higher downstream model quality.}
  \label{tab:dual_bin_retention_cc51}
\end{table}

$\phi_{\mathrm{DCLM}}$ preferentially selects documents with high world-knowledge density; $\phi_{\mathrm{BETR}}$ preferentially selects documents with strong benchmark-proximate reasoning signal. Their union produces a training corpus with a more balanced and task-aligned capability profile than either single-bin policy is capable of achieving, as demonstrated empirically in Section~\ref{sec:experiments}.Full threshold sweep statistics including per-configuration retention and per-benchmark accuracy are provided in Appendix~\ref{sec:classifier_ablations}.

\subsection{The Final FLUX Pipeline}
To establish the final FLUX configuration, we perform a staged ablation over parser selection, heuristic filtering, deduplication, and classification. We conducted multiple ablations across model sizes ranging from \textbf{530M to 1B parameters} to systematically evaluate the impact of different components in the pipeline and identify the most effective configuration at each stage. On the \textbf{530M-parameter model}, the \textbf{aggregate score improves progressively from 41.23 to 42.43 to 45.45 to 48.22} as each stage from Parser followed by Filters, Deduplication and Classification respectively. This trend is further reflected in the \textbf{3B model evaluation}, where the final pipeline achieves an \textbf{aggregate score of 51.92\% and an MMLU score of 32.14\%}, demonstrating consistent gains across model scales.

\begin{figure}[H]
\centering
\includegraphics[width=0.92\linewidth]{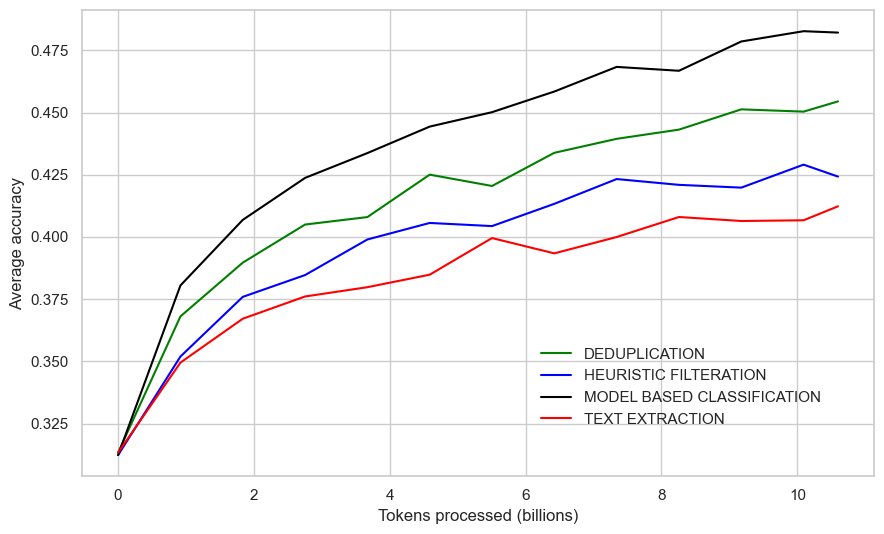}
\caption{\textbf{Progressive improvement across FLUX pipeline stages (530M scale; 10.6B tokens).} Each successive pipeline stage --- text extraction, heuristic filtering, deduplication, and model based classification --- yields consistent aggregate score gains, confirming that all four components contribute meaningfully to final dataset quality.}
\label{fig:flux_pipeline_progression}
\end{figure}

\section{Decontamination}\label{sec:decontamination}
To mitigate train--test contamination~\citep{jacovi_stop_2023,lee_deduplicating_2022}, benchmark-derived text was removed from the raw pretraining pool before any optimization step. We use the \texttt{decon} pipeline~\citep{allenai2024decon}, which performs n-gram overlap screening to identify exact and near-exact lexical matches between candidate corpus documents and evaluation references.

The decontamination pass was executed over approximately 62M documents from the FLUX corpus (\textasciitilde60B tokens). Each document was matched against a unified benchmark reference set comprising MMLU, HellaSwag, ARC-Easy, ARC-Challenge, CSQA, PIQA, SocialIQA, WinoGrande, and OpenBookQA~\citep{hendrycks_measuring_2021,zellers_hellaswag_2019,clark_think_2018,talmor_commonsenseqa_2019,bisk_piqa_2019,sap_socialiqa_2019,sakaguchi_winogrande_2019,mihaylov_can_2018}. Documents with similarity scores above the configured threshold (exact or near-exact matches) were deterministically marked for removal and excluded at corpus scope before downstream preprocessing stages (parsing, filtering, deduplication, and quality classification). The same exclusion policy was enforced for all pretraining runs, ensuring contamination-free inputs across model scales from 530M to 3B parameters.

Table~\ref{tab:decon_stats} reports the contamination statistics. A total of 243 unique documents exhibited overlap with evaluation benchmarks, corresponding to 517 contaminated evaluation instances. The overall contamination rate (243 / 62M $\approx 3.9 \times 10^{-6}$) was negligible relative to corpus size. MMLU exhibited the highest contamination (117 documents, 322 instances), while SocialIQA and OpenBookQA each contained a single contaminated document.

\begin{table}[H]
  \centering
  \small
  \setlength{\tabcolsep}{8pt}
  \renewcommand{\arraystretch}{1.15}
  \begin{tabular}{l|c|c}
    \hline
    \textbf{Benchmark} & \textbf{Unique Contaminated Documents} & \textbf{Contaminated Evaluation Instances} \\
    \hline
    MMLU & 117 & 322 \\
    HellaSwag & 73 & 82 \\
    ARC-Challenge & 23 & 29 \\
    ARC-Easy & 21 & 37 \\
    CSQA & 9 & 34 \\
    PIQA & 7 & 8 \\
    SocialIQA & 1 & 1 \\
    Winogrande & 3 & 3 \\
    OpenBookQA & 1 & 1 \\
    \hline
    \textbf{Total} & \textbf{243} & \textbf{517} \\
    \hline
  \end{tabular}
  \caption{\textbf{Decontamination statistics} against evaluation benchmarks.}
  \label{tab:decon_stats}
\end{table}
\section{Experiments}
\label{sec:experiments}
\noindent We evaluate FLUX against DCLM and FineWeb across multiple model scales using the nine-benchmark suite described in Appendix~\ref{app:eval_suite}~\citep{li_datacomp-lm_2025,penedo_fineweb_2024}. All results are reported after benchmark decontamination of the pretraining corpus (Section~\ref{sec:decontamination}). Per-benchmark learning curves at 3B scale and results at 530M and 1B scales are provided in Appendix~\ref{app:extended_eval}.

\subsection{FLUX vs. DCLM}

Aggregate learning-curve analysis (3B scale; 60B tokens) shows that FLUX consistently outperforms both DCLM and FineWeb across the full training token budget, with the gap over DCLM widening steadily beyond 30B tokens. Table~\ref{tab:3b_results} reports final checkpoint scores. FLUX achieves an aggregate score of 51.92 across the nine-benchmark suite, outperforming DCLM (50.48) by 1.44 percentage points and FineWeb (48.83) by 3.09 percentage points. On MMLU, FLUX scores 32.14\% against DCLM's 31.98\% and FineWeb's 29.88\%. FLUX outperforms DCLM on 7 of 9 benchmarks, with the largest gains on ARC-Challenge (+2.9pp), WinoGrande (+2.9pp), SocialIQA (+2.2pp), and PIQA (+1.9pp). At the corpus level, FLUX extracts 50B tokens from a single Common Crawl dump versus 40B for DCLM, a 25\% improvement in token retention, and yields 365B tokens versus 302B post-deduplication across two dumps, a 21\% gain. These results demonstrate that the retention advantage of FLUX does not trade off against downstream quality — FLUX strictly dominates DCLM on both axes simultaneously.

\begin{table}[H]
  \centering
  \small
  \setlength{\tabcolsep}{4pt}
  \renewcommand{\arraystretch}{1.1}
  \resizebox{\linewidth}{!}{%
    \begin{tabular}{lcccccccccc}
      \toprule
      \textbf{Pipeline} & \textbf{MMLU} & \textbf{ARC-Easy} & \textbf{ARC-Challenge} & \textbf{CSQA} & \textbf{HellaSwag} & \textbf{OpenBookQA} & \textbf{PIQA} & \textbf{SocialIQA} & \textbf{WinoGrande} & \textbf{Aggregate} \\
      \midrule
      FineWeb & 0.298 & 0.602 & 0.337 & 0.548 & 0.515 & 0.400 & 0.706 & 0.471 & 0.518 & 0.4883 \\
      DCLM & 0.319 & \textbf{0.673} & 0.343 & 0.548 & 0.524 & \textbf{0.454} & 0.703 & 0.467 & 0.513 & 0.5048 \\
      \rowcolor{blue!10} FLUX & \textbf{0.321} & 0.666 & \textbf{0.372} & \textbf{0.570} & \textbf{0.541} & 0.450 & \textbf{0.722} & \textbf{0.489} & \textbf{0.542} & \textbf{0.5192} \\
      \bottomrule
    \end{tabular}%
  }
  \caption{\textbf{Performance across the nine-benchmark evaluation suite (3B model, 60B training tokens).} FLUX achieves the highest aggregate score (51.92) compared to DCLM (50.48) and FineWeb (48.83). FLUX also obtains the best results on 7 of the 9 benchmarks in the evaluation suite.}
  \label{tab:3b_results}
\end{table}

\begin{figure}[H]
  \centering
  \includegraphics[width=0.90\linewidth]{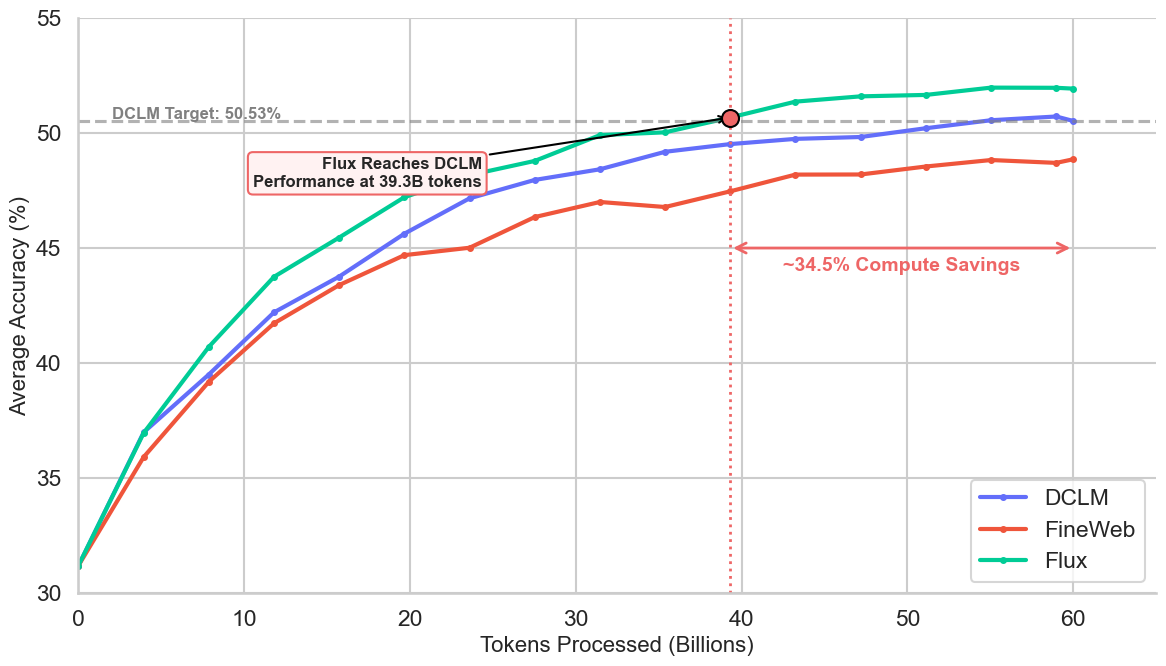}
  \caption{\textbf{Final Comparison Overview (3B scale; 60B tokens).} Aggregate learning-curve comparison across FLUX, DCLM, and FineWeb under a matched token budget.}
  \label{fig:3b_final_overview}
\end{figure}

\subsection{FLUX-Base vs. FineWeb}

To isolate the contribution of the upstream pipeline from the classifier stage, we evaluate
\textbf{FLUX-Base} --- the FLUX pipeline with the dual-bin classification stage removed --- against
FineWeb in a controlled quality-and-retention comparison. This configuration represents the most
direct test of whether FLUX's filtering and deduplication design alone is competitive, independent
of any post-hoc model-based selection.

\begin{figure}[H]
  \centering
  \includegraphics[width=0.90\linewidth,trim=12 10 12 10,clip]{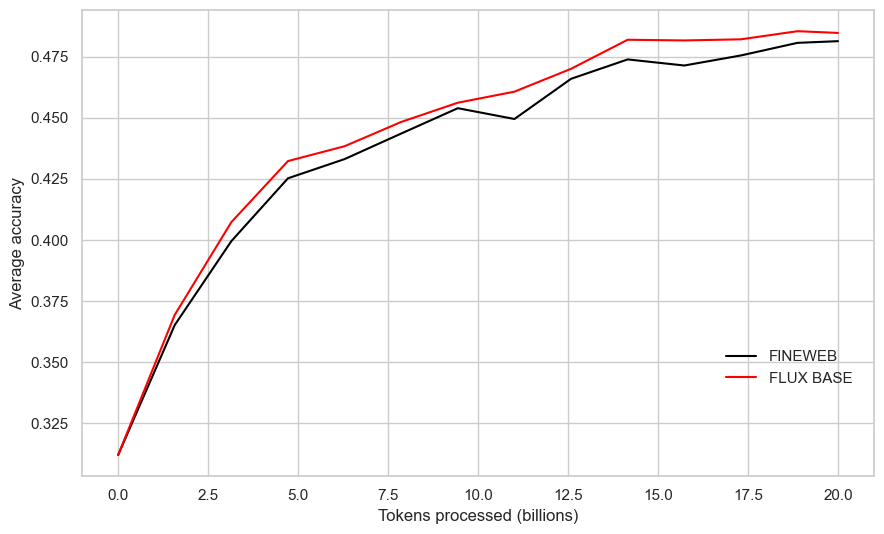}
  \caption{\textbf{FLUX-Base vs FineWeb (1B scale; 20B tokens).} Comparison between FLUX-Base and FineWeb, emphasizing quality-retention behavior when FLUX is evaluated without the final classifier stage.}
  \label{fig:flux_base_fineweb}
\end{figure}

\paragraph{Downstream quality.} Figure~\ref{fig:flux_base_fineweb} reports aggregate learning
curves at the (1B scale; 20B tokens). FLUX-Base outperforms FineWeb consistently across the full
training token budget. Table~\ref{tab:1b_results} reports final checkpoint scores: FLUX-Base
achieves an aggregate score of 48.53 against FineWeb's 48.05, leading on 5 of 9 benchmarks, with
the largest margins on CSQA (+3.4~pp) and ARC-Challenge (+3.0~pp).

\paragraph{Token retention.} At the corpus level, FLUX-Base yields 192B tokens from the same raw
input (CC-MAIN-2025-51) against which FineWeb produces 170B tokens --- a 12.9\% improvement in retention.
Importantly, this advantage is achieved \emph{with} comprehensive cross-dump deduplication:
FLUX-Base applies Bloom Filter-based fuzzy deduplication globally across both Common Crawl dumps
(Appendix~\ref{app:dedup}), whereas FineWeb foregoes global deduplication to reach its scale.
FLUX-Base therefore achieves both higher retention and stronger deduplication simultaneously.

These results establish that FLUX's gains do not depend on the classifier stage. The upstream
filtering and deduplication pipeline alone is sufficient to exceed FineWeb on both token
retention and downstream model quality.

\begin{table}[H]
  \centering
  \small
  \setlength{\tabcolsep}{4pt}
  \renewcommand{\arraystretch}{1.1}
  \resizebox{\linewidth}{!}{%
    \begin{tabular}{l S[table-format=1.3] S[table-format=1.3] S[table-format=1.3] S[table-format=1.3] S[table-format=1.3] S[table-format=1.3] S[table-format=1.3] S[table-format=1.3] S[table-format=1.3] S[table-format=1.3]}
      \toprule
      \textbf{Pipeline} & \textbf{MMLU} & \textbf{ARC-Easy} & \textbf{ARC-Challenge} & \textbf{CSQA} & \textbf{HellaSwag} & \textbf{OpenBookQA} & \textbf{PIQA} & \textbf{SocialIQA} & \textbf{WinoGrande} & \textbf{Aggregate} \\
      \midrule
      FineWeb & 0.294729 & {\bfseries\num{0.599}} & 0.313993 & 0.503686 & {\bfseries\num{0.524}} & {\bfseries\num{0.386}} & 0.691 & 0.481 & {\bfseries\num{0.534333}} & 0.481415667 \\
      \rowcolor{blue!10} FLUX-Base & {\bfseries\num{0.297655}} & 0.596 & {\bfseries\num{0.314846}} & {\bfseries\num{0.542998}} & 0.522 & 0.380 & {\bfseries\num{0.703}} & {\bfseries\num{0.486}} & 0.525651 & {\bfseries\num{0.484794444}} \\
      \bottomrule
    \end{tabular}%
  }
  \caption{\textbf{Comparison of FLUX-Base and FineWeb across the nine-benchmark evaluation suite (1B model, 20B training tokens).} FLUX-Base achieves a slightly higher aggregate score (48.53) than FineWeb (48.05), and outperforms FineWeb on 5 of the 9 benchmarks.}
  \label{tab:1b_results}
\end{table}

\newpage

\section{Conclusion and Limitations}
In this work, we introduced FLUX, a web-scale data preprocessing pipeline designed to systematically overcome the long-standing retention--quality trade-off in language model pretraining. Unlike prior approaches such as DCLM and FineWeb, which optimize along a single axis, FLUX demonstrates that it is possible to jointly maximize token retention while improving downstream model performance.

Empirically, models trained on FLUX consistently outperform strong baselines across all evaluation settings. At the 3B scale, FLUX achieves 32.14\% MMLU accuracy, exceeding DCLM (31.98\%) and FineWeb (29.88\%), while also improving the aggregate score across the full nine-benchmark suite. At the data level, FLUX extracts 50B usable tokens from a single dump (CC-MAIN-2025-51), compared to 40B from DCLM (+25\% retention), and scales up to 192B tokens without classification, surpassing FineWeb's 170B while maintaining stronger quality. These results establish that retention, when paired with targeted quality controls, directly translates into better model capability.

The improvements are driven by three key design principles. First, FLUX replaces aggressive document-level rejection with line-level cleaning, recovering substantial amounts of usable signal. Second, it adopts whole-document language identification, avoiding unnecessary loss of high-quality English content. Third, the dual-bin classification framework leverages complementary signals for knowledge and reasoning, producing a more balanced and task-aligned training corpus. Together, these components yield a dataset that is not only larger but also denser, cleaner, and more effective for training.

Despite these gains, several limitations remain. FLUX is evaluated under an English-only setting, and its effectiveness for multilingual corpus construction is not yet established. Deduplication is performed across only two Common Crawl snapshots (CC-MAIN-2025-47 and CC-MAIN-2025-51), and extending global deduplication across the complete archive of 128 historical dumps remains an open challenge. Additionally, all experiments are conducted up to the 3B parameter scale, leaving the interaction between data curation and larger model regimes unexplored. Finally, while the evaluation suite is comprehensive, it remains benchmark-driven and may not fully capture real-world deployment behavior.

Overall, FLUX establishes a new operating point for web-scale dataset construction: higher retention, stronger quality, and better downstream performance, simultaneously. We believe this work provides a foundation for future data-centric research, particularly in extending these principles to multilingual settings, larger model scales, and more diverse evaluation regimes.

\newpage

\appendix
\newpage
\section*{Appendix}
\appendixtoc
\newpage
\appsection{Training Setup and Hyperparameters}
\label{app:training_details}
\paragraph{Overview.}
Our training setup follows closely that of Groeneveld et al.~\citep{groeneveld_olmo_2024}. Specifically, we build our training infrastructure using the OLMo~\citep{groeneveld_olmo_2024} codebase, which supports decoder-only, pre-normalization Transformers following GPT-style architectures~\citep{radford_improving_nodate,touvron_llama_2023}. OLMo is implemented in PyTorch~\citep{paszke_pytorch_2019} and targets Distributed Data Parallel (DDP) training~\citep{li_pytorch_2020}. All experiments are conducted in bf16 precision~\citep{micikevicius_mixed_2018}. We train models on the FLUX corpus, and conduct controlled comparisons against models trained on DCLM and FineWeb under identical training settings~\citep{li_datacomp-lm_2025,penedo_fineweb_2024}.

\paragraph{Reproducibility.}
Unless stated otherwise, all comparisons use identical training recipes (sequence length, batch size, and optimizer hyperparameters) and token budgets across model scales (530M, 1B, and 3B). We report single-run results per scale using a fixed random seed for initialization and data shuffling; additional multi-seed runs and variance estimates will be added in later revisions.

\paragraph{Architecture details.}
We utilize RMSNorm~\citep{zhang2019rmsnorm} for all normalization layers without bias parameters and SwiGLU~\citep{shazeer2020glu} multilayer perceptrons (MLPs) with an expansion ratio of 8. All linear layers exclude bias terms. We adopt Rotary Positional Embeddings (RoPE)~\citep{su_roformer_2023} with $\theta = 10{,}000$ and enable full-precision rotary computation. FlashAttention~\citep{dao_flashattention_2022} is used for efficient attention computation. Our sequence length during pretraining is 2{,}048. We pack multiple sequences into batches to fill the entire context, with an EOS token used to split documents~\citep{raffel_exploring_2023,groeneveld_olmo_2024}. We allow causal attention to attend across document boundaries. Parameters are initialized from a truncated normal distribution with standard deviation 0.02~\citep{glorot_understanding_2010}. Weight tying is disabled.

\paragraph{Training sets and tokenization.}
Since the focus of our paper is dataset development, we train primarily on the FLUX corpus, constructed from filtered web-scale data. For ablation studies, we additionally train models on DCLM and FineWeb using the same model architecture and optimization settings to enable controlled comparisons. For all experiments we use the EleutherAI GPT-NeoX-20B-PII-Special tokenizer, yielding a vocabulary size of approximately 50k~\citep{gptneox2022}. Data loading uses right padding and drops incomplete batches. We apply a repetition-based instance filter to remove locally repetitive spans prior to sequence packing.

\paragraph{Optimization details.}
As mentioned in the main body, we train with a standard next-token prediction objective~\citep{radford_improving_nodate}. We employ the AdamW optimizer~\citep{loshchilov2019adamw} with $\beta_1 = 0.9$, $\beta_2 = 0.95$, and weight decay 0.1. Gradients are clipped to a maximum norm of 1.0~\citep{pascanu_difficulty_2013}. We use a cosine learning rate schedule with linear warmup and decay the learning rate to a fixed fraction of its peak value~\citep{loshchilov_sgdr_2017}.

\paragraph{Hyperparameters.}
We detail the hyperparameters for our models in Table~\ref{tab:training_hparams}. For the 530M and 1B scales, we use learning rates tuned for stable perplexity optimization. For the 3B scale, we adopt a lower peak learning rate and longer warmup, together with a smaller final decay factor. Weight decay is fixed to 0.1 across all experiments. All comparisons between FLUX, DCLM, and FineWeb are conducted with identical architectural and optimization hyperparameters.

\begin{table}[H]
  \centering
  \small
  \setlength{\tabcolsep}{5pt}
  \renewcommand{\arraystretch}{1.15}
  \begin{tabular}{
    >{\raggedright\arraybackslash}p{1.0cm}
    >{\centering\arraybackslash}p{1.0cm}
    >{\centering\arraybackslash}p{1.0cm}
    >{\centering\arraybackslash}p{1.6cm}
    >{\centering\arraybackslash}p{1.0cm}
    >{\centering\arraybackslash}p{2.0cm}
    >{\centering\arraybackslash}p{1.3cm}
    >{\centering\arraybackslash}p{2.0cm}
    >{\centering\arraybackslash}p{1.2cm}
  }
    \toprule
    \textbf{Scale} &
    \textbf{Layers (L)} &
    \textbf{Heads (H)} &
    \textbf{$d_{model}$} &
    \textbf{$d_{head}$} &
    \textbf{Peak LR} &
    \textbf{Weight decay} &
    \textbf{Final LR frac ($\alpha_f$)} &
    \textbf{Global batch} \\
    \midrule
    530M & 16 & 16 & 1{,}344 & 84 & $2.8\times 10^{-3}$ & 0.1 & 0.1 & 448 \\
    1B & 16 & 16 & 2{,}048 & 128 & $2.1\times 10^{-3}$ & 0.1 & 0.1 & 768 \\
    3B & 24 & 16 & 2{,}752 & 172 & $1.0\times 10^{-4}$ & 0.1 & 0.05 & 960 \\
    \bottomrule
  \end{tabular}
  \caption{\textbf{Model scales and training hyperparameters.} For each model scale, we report the number of transformer layers (L), attention heads (H), hidden dimension ($d_{model}$), per-head dimension ($d_{head}$), peak learning rate, weight decay, final learning rate fraction ($\alpha_f$), and global batch size. Batch sizes are reported in sequences of length 2{,}048 tokens.}
  \label{tab:training_hparams}
\end{table}

\appsection{Evaluation Suite and Metrics}
\label{app:eval_suite}
\label{sec:eval_suite}
We evaluate with the \texttt{lighteval} framework~\citep{lighteval2023} in a 5-shot setting to reduce prompt sensitivity and emphasize in-context reasoning. All metrics are computed via log-likelihood ranking over candidate answers. Our evaluation suite covers nine benchmarks:
\begin{itemize}
  \item \textbf{MMLU (Massive Multitask Language Understanding):} breadth across 57 STEM and humanities subjects~\citep{hendrycks_measuring_2021}.
  \item \textbf{ARC-Easy \& ARC-Challenge:} grade-school science reasoning with increasing difficulty~\citep{clark_think_2018}.
  \item \textbf{CSQA (CommonsenseQA):} commonsense reasoning over everyday scenarios~\citep{talmor_commonsenseqa_2019}.
  \item \textbf{HellaSwag:} commonsense inference via plausible narrative continuation~\citep{zellers_hellaswag_2019}.
  \item \textbf{OpenBookQA:} multi-step science QA combining provided facts with external knowledge~\citep{mihaylov_can_2018}.
  \item \textbf{PIQA (Physical Interaction QA):} physical commonsense about object affordances~\citep{bisk_piqa_2019}.
  \item \textbf{SocialIQA:} reasoning about social interactions, motives, and emotions~\citep{sap_socialiqa_2019}.
  \item \textbf{WinoGrande:} adversarial pronoun resolution to test bias-resistant commonsense~\citep{sakaguchi_winogrande_2019}.
\end{itemize}
This 5-shot, nine-benchmark protocol underlies the performance comparisons reported in Section~\ref{sec:experiments}.

\appsection{Text Extraction: Parser Design and Token Yield Analysis}
\label{app:text_extraction}
\label{sec:text_extraction}

\appsubsection{Overview}
Web-scale text extraction pipelines inherently face a critical trade-off between extraction quality and computational efficiency~\citep{barbaresi_trafilatura_2021,bevendorff_elastic_chatnoir_2021,li_datacomp-lm_2025}. High-fidelity pipelines consistently deliver superior output but demand considerable computational resources, while lightweight alternatives often introduce noise, encoding inconsistencies, and unreliable boilerplate removal~\citep{penedo_fineweb_2024,soldaini_dolma_2024}. To systematically investigate this trade-off, we design and evaluate four distinct parsing pipelines Parser V1, Parser V2, Apex, and Parser V4 each varying in its approach to content boundary detection, boilerplate filtering, and encoding normalization. The primary objective is to identify the optimal configuration that maximizes output quality while maintaining computational feasibility.

To rigorously measure the practical impact of these design decisions, each parser's output is used as the training corpus for a 530M-parameter language model, trained under strictly controlled and identical experimental conditions. This methodology ensures that parser quality remains the sole variable under examination. Performance is assessed using effective training token yield — the total volume of clean, non-redundant, and semantically coherent text produced by each parser. This evaluation is conducted independently across both multilingual and English-only datasets, ensuring a thorough and unbiased comparative analysis.

\appsubsection{Token Yield Across Parsers}
Table~\ref{tab:parser_token_yield} reports total token counts, expressed in billions, produced by each parser across both content configurations on a standardized evaluation set of 10K WARC files from CC-MAIN-2025-51 snapshot.

\begin{table}[H]
  \centering
  \small
  \setlength{\tabcolsep}{8pt}
  \renewcommand{\arraystretch}{1.15}
  \begin{tabular}{lcc}
    \toprule
    \textbf{Parser} & \textbf{Multilingual (B tokens)} & \textbf{English (B tokens)} \\
    \midrule
    Parser V1 & 391.3 & 50.7 \\
    Parser V2 & 327.4 & 49.6 \\
    Parser V3 & 153.4 & 41.0 \\
   \rowcolor{blue!10} Apex  & 504.5 & 57.9 \\
    
    \bottomrule
  \end{tabular}
  \caption{\textbf{Token yield per parser across multilingual and English corpora.}}
  \label{tab:parser_token_yield}
\end{table}

As reported in Table~\ref{tab:parser_token_yield}, Apex delivers the highest token yield across both evaluation settings, with 504.5B multilingual tokens and 57.9B English tokens. By comparison, Parser V1 yields 391.3B/50.7B and Parser V2 yields 327.4B/49.6B, whereas Parser V3 is markedly more conservative at 153.4B multilingual and 41.0B English tokens. Notably, the wider spread observed in the multilingual regime indicates that encoding robustness and language-agnostic boilerplate suppression are the principal sources of performance separation across parser designs. Taken together, these results support selecting Apex as the production parser for the FLUX pipeline.

\appsubsection{Parser Architecture Comparison}
Table~\ref{tab:parser_architecture} presents a structured comparison of the parser implementations across four core architectural dimensions: WARC ingestion, HTML parsing backend, content extraction strategy, and encoding detection methodology.

\begin{table}[H]
  \centering
  \small
  \setlength{\tabcolsep}{4pt}
  \renewcommand{\arraystretch}{1.1}
  \resizebox{\linewidth}{!}{%
    \begin{tabular}{lcccc}
      \toprule
      \textbf{Component} & \textbf{Parser V1} & \textbf{Parser V2} & \textbf{Parser V3} & \textbf{Apex} \\
      \midrule
      WARC Reader & Custom streaming & Custom streaming & FastWARC (C) & FastWARC (C) \\
      HTML Parser & Selectolax/Lexbor & Selectolax/Lexbor & Resiliparse & Resiliparse \\
      Extraction Strategy & Container-first & Container-first & Element-wise traversal & Statistical main-content  \\
      Encoding Detection & Multi-stage heuristic & Multi-stage heuristic & Statistical inference & C-level detection  \\
      \bottomrule
    \end{tabular}%
  }
  \caption{\textbf{Architectural comparison of the four parser implementations.} The four parsers differ primarily in their content extraction strategy --- from container-first (V1/V2) and element-wise traversal (V3) to Apex's statistical main-content detection --- which is the principal driver of performance and yield differences.}
  \label{tab:parser_architecture}
\end{table}

\appsubsection{Parser V1 and V2: Container-First Extraction}
Parser V1 and Parser V2 are built upon a shared architectural foundation, both leveraging a custom WARC streaming reader that affords precise control over record boundary detection and HTTP response segmentation. HTML parsing is delegated to Selectolax~\citep{selectolax_github}, driven by the Lexbor engine~\citep{selectolax_github} a lightweight, C-implemented parser engineered for high tokenization throughput making it especially effective at handling the syntactically irregular markup characteristic of large-scale web crawls.

Both parsers adopt a container-first extraction strategy, wherein targeted DOM traversal is performed to isolate high-density content regions. These regions encompass semantic elements such as div, article, and main, as well as div-based wrappers exhibiting elevated text-to-markup ratios a design choice that substantially reduces the inadvertent inclusion of boilerplate content. Encoding normalization is governed by a structured, multi-stage heuristic pipeline that sequentially interrogates HTTP headers, HTML meta charset declarations, and byte-order marks, falling back to statistical byte-frequency analysis only when upstream signals prove inconclusive.

Parser V2 builds upon Parser V1 through a set of targeted refinements in content boundary detection and noise filtering, enhancing the system's capacity to discriminate between primary editorial content and peripheral navigational or promotional elements, without altering the underlying processing architecture. The observed token yield differential 391.3B tokens for Parser V1 against 327.4B tokens for Parser V2 across the multilingual corpus is a direct consequence of these more stringent filtering criteria. This quantitative gap substantiates the conclusion that Parser V2 occupies a higher precision, lower recall operating point relative to its predecessor, trading corpus volume for compositional integrity.

\appsubsection{Parser V3: Element-Wise Extraction}
Parser V3 retains an established parsing stack comprising FastWARC for archive ingestion and Resiliparse for HTML parsing, while introducing a fundamentally distinct extraction methodology. In contrast to approaches operating at container granularity or employing statistical region scoring, Parser V3 performs element-wise DOM traversal with per-element retention logic conditioned on tag semantics, positional context, and local content heuristics. Inline elements are assessed by their textual contribution relative to immediate siblings, while block-level elements are evaluated on content-to-noise ratio within their respective subtrees prior to inclusion.

This fine-grained extraction strategy affords precise control over content boundaries, enabling selective retention from structurally heterogeneous documents where coarser methods are susceptible to discarding semantically valid text regions. The resulting token yield 153.4B multilingual tokens and 41.0B English tokens is more plausibly attributable to enhanced extraction precision than to diminished document coverage, reflecting the stricter element-level filtering inherent to this design. Parser V3 thus occupies a high-precision, lower-recall position within the parser design space and is retained as a reference configuration for ablation studies prioritizing extraction fidelity over throughput.

\appsubsection{Apex: Statistical Main-Content Detection}
Apex represents a substantive architectural advancement over Parser V1 and Parser V2, introducing component-level replacements across the entire processing stack. WARC ingestion is handled by FastWARC, a C-implemented archive reader offering lower per-record latency and reduced memory overhead over custom streaming implementations. HTML parsing is delegated to Resiliparse a robust extraction framework purpose-built for large-scale web data processing  capable of accommodating malformed markup, tag soup, and encoding inconsistencies endemic to real-world crawl corpora.

Rather than relying on fixed semantic tag sets, Apex employs statistical main-content detection, evaluating candidate page regions through a composite scoring function over three signals: content density (text-to-node character ratio within a subtree), link density (proportion of anchor-enclosed text within a region), and structural depth relative to the document root. Regions satisfying empirically calibrated thresholds across all three signals are retained as primary content; remaining nodes are discarded as boilerplate. This signal-driven approach enables reliable extraction from documents with non-standard structural layouts, where container-first strategies typically underperform.

Apex achieves the highest token yield across both evaluation configurations, producing 504.5B multilingual tokens and 57.9B English tokens, and is accordingly adopted as the production parser for the FLUX pipeline.

\begin{figure}[H]
  \centering
  \includegraphics[width=\linewidth]{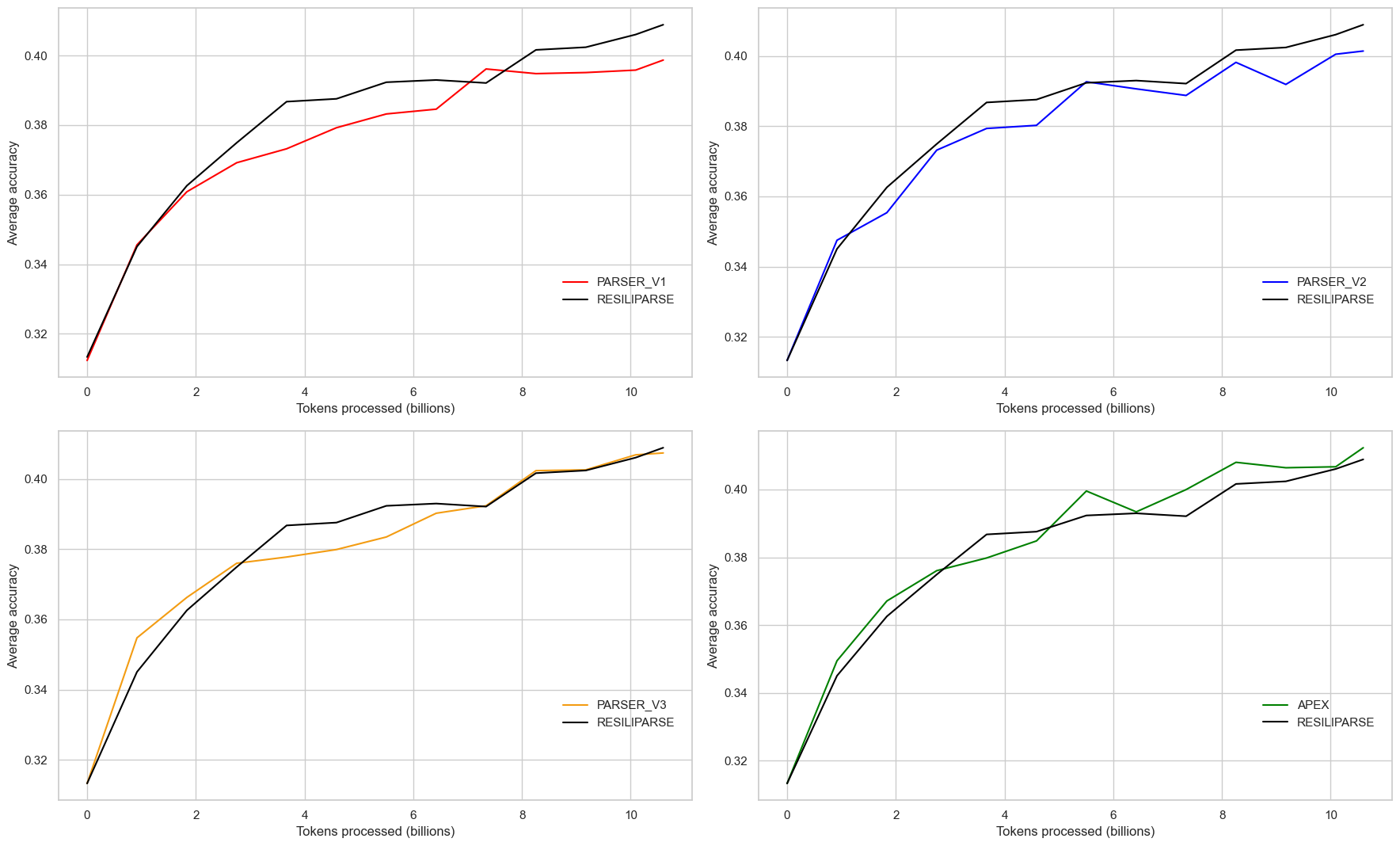}
  \caption{\textbf{Parser-stage comparison across configurations (530m scale; 10.6B tokens).} Combined view of parser outputs used in the parser analysis.}
  \label{fig:parser_combined_grid}
\end{figure}

\appsection{Filtering Pipeline: Stage-by-Stage Specification}
\label{app:filter_pipeline}

\appsubsection{Filter Specifications and Thresholds}
\label{sec:filter_pipeline_details}
The FLUX filtering subsystem is a deterministic 13-stage pipeline operating over pre-parsed JSONL documents derived from Common Crawl WARC archives. Stages are partitioned into four functional groups: URL-based pre-filtering and normalization (Stages 1--6), language identification (Stage 7), document-level quality gating (Stages 8--11), and line-level content cleaning with integrity enforcement (Stages 12--13). Rejection at any stage is unconditional. The complete stage listing is given in Table~\ref{tab:d1_stage_overview} and per-filter removal statistics are reported in Appendix~Section~\ref{app:filter_d2}.

A governing design principle is the preference for intra-document line-level excision over document-level rejection wherever noise is localized: Stage 12 excises individual boilerplate lines and retains the cleaned document body, and Stage 13 enforces a post-cleaning integrity constraint rejecting documents for which excision removes more than 5\% of pre-cleaning tokens.

\begin{table}[H]
  \centering
  \small
  \setlength{\tabcolsep}{6pt}
  \renewcommand{\arraystretch}{1.15}
  \begin{tabular}{clll}
    \toprule
    \textbf{\#} & \textbf{Component} & \textbf{Functional Group} & \textbf{Type} \\
    \midrule
    1 & Document Parsing & URL \& Normalization & Structural preprocessor \\
    2 & UT1 Domain Blocklist & URL \& Normalization & Document filter \\
    3 & URL Strict Substring & URL \& Normalization & Document filter \\
    4 & URL Hard Substring & URL \& Normalization & Document filter \\
    5 & URL Soft Substring & URL \& Normalization & Document filter \\
    6 & URL Token Removal \& Newline Norm. & URL \& Normalization & Content modifier (no rejection) \\
    7 & Language Identification & Language ID & Document filter \\
    8 & Gopher Quality & Document-level Quality & Document filter \\
    9 & Nemo & Document-level Quality & Document filter \\
    10 & Gopher Repetition & Document-level Quality & Document filter \\
    11 & Custom Quality Filter & Document-level Quality & Document filter \\
    12 & Line-level Cleaning & Line-level \& Integrity & Content modifier + conditional rejection \\
    13 & Word Removal Ratio & Line-level \& Integrity & Document filter \\
    \bottomrule
  \end{tabular}
  \caption{\textbf{FLUX filtering pipeline: 13 stages in execution order.}}
  \label{tab:d1_stage_overview}
\end{table}

\subsubsection{Stages 1--6: URL-based Pre-filtering and Normalization}
\paragraph{Stage 1: Document Parsing}
Each input record is deserialized from its JSONL representation. The canonical URL is extracted from the WARC \texttt{Target-URI} field (fallback: \texttt{url} field). Records failing deserialization are excluded and are not counted in any rejection statistic.

\paragraph{Stage 2: UT1 Domain Blocklist Filter}
The registered domain of each document URL is matched by exact lookup against the UT1 blocklist~\citep{ut1_blocklist}. Any domain appearing in one or more blocklist categories triggers unconditional rejection prior to any content-based analysis.

\paragraph{Stage 3: URL Strict Substring Filter}
The document URL is evaluated against a high-precision bad-word lexicon. A match requires the candidate substring to occur as a token-delimited unit within the URL path, using \texttt{-} and \texttt{.} as delimiters. A single confirmed match triggers unconditional rejection.

\paragraph{Stage 4: URL Hard Substring Filter}
The document URL is evaluated against a secondary bad-word lexicon with no delimiter constraint. Any occurrence of a hard substring anywhere in the full URL string triggers rejection on a single match.

\paragraph{Stage 5: URL Soft Substring Filter}
The document URL is evaluated against a tertiary bad-word lexicon requiring a minimum co-occurrence of two matching substrings before rejection is triggered.

\paragraph{Stage 6: URL Token Removal and Newline Normalization}
Two deterministic content modifiers are applied sequentially; neither triggers document rejection. URL token removal strips inline URL strings via a TLD-anchored pattern matcher. Newline normalization collapses runs of $\geq 3$ consecutive newlines to \texttt{\textbackslash n\textbackslash n}.

\begin{table}[H]
  \centering
  \small
  \setlength{\tabcolsep}{10pt}
  \renewcommand{\arraystretch}{1.15}
  \begin{tabular}{lrrr}
    \toprule
    \textbf{Filter (Stage)} & \textbf{Docs Removed} & \textbf{Tokens Removed} & \textbf{\% Total Tokens} \\
    \midrule
    UT1 Domain Blocklist (2) & 12{,}057 & 16{,}752{,}457 & 0.26\% \\
    URL Strict Substring (3) & 343 & 514{,}779 & 0.01\% \\
    URL Hard Substring (4) & 2{,}545 & 3{,}723{,}647 & 0.06\% \\
    URL Soft Substring (5) & 396 & 619{,}897 & 0.01\% \\
    URL Token Removal\textsuperscript{$\dagger$} (6) & --- & 73{,}019{,}753 & 1.13\% \\
    \bottomrule
  \end{tabular}
  \caption{\textbf{URL-stage removal statistics (10{,}000 WARC files from CC-MAIN-2025-13 snapshot).} \textsuperscript{$\dagger$}Content modifier; no document rejection.}
  \label{tab:d1_url_stats}
\end{table}

\subsubsection{Stage 7: Language Identification}
\paragraph{Stage 7: Language Identification}
Each document is scored over 176 language classes by the FastText \texttt{lid.176} model, applied to normalized text. A document is admitted to the English corpus iff its English-class confidence score is at least 0.65. Documents below threshold are routed to a separate multilingual partition and are not discarded. The LID strategy ablation and this operating-point choice are reported in Appendix~\ref{app:lid_ablation} (D.4).

\subsubsection{Stages 8--11: Document-level Quality Gating}
\paragraph{Stage 8: Gopher Quality Filter}
Seven statistical quality criteria are evaluated with early exit on first failure. A document is rejected if any single criterion is violated.

\begin{table}[H]
  \centering
  \small
  \setlength{\tabcolsep}{10pt}
  \renewcommand{\arraystretch}{1.15}
  \begin{tabular}{lcc}
    \toprule
    \textbf{Criterion} & \textbf{Min Threshold} & \textbf{Max Threshold} \\
    \midrule
    Document word count & 50 & 100{,}000 \\
    Mean word length (chars) & 3.0 & 10.0 \\
    Symbol-to-word ratio & --- & 0.10 \\
    Bullet-line ratio & --- & 0.90 \\
    Ellipsis-line ratio & --- & 0.30 \\
    Alphabetic word ratio & 0.80 & --- \\
    Stop-word count & 2 & --- \\
    \bottomrule
  \end{tabular}
  \caption{\textbf{Gopher Quality Filter thresholds.}}
  \label{tab:d1_gopher_thresholds}
\end{table}

\paragraph{Stage 9: Nemo Filter}
Five independent character-ratio constraints are applied; a document is rejected if any single constraint is violated.

\begin{table}[H]
  \centering
  \small
  \setlength{\tabcolsep}{10pt}
  \renewcommand{\arraystretch}{1.15}
  \begin{tabular}{lc}
    \toprule
    \textbf{Criterion} & \textbf{Max Threshold} \\
    \midrule
    Non-alphanumeric character ratio & 0.25 \\
    Numeric character ratio & 0.15 \\
    URL character ratio & 0.20 \\
    Whitespace character ratio & 0.25 \\
    Parenthesis character ratio & 0.10 \\
    \bottomrule
  \end{tabular}
  \caption{\textbf{Nemo Filter thresholds.}}
  \label{tab:d1_nemo_thresholds}
\end{table}

\paragraph{Stage 10: Gopher Repetition Filter}
Thirteen repetition metrics are computed across line, paragraph, and word n-gram granularities. A document is rejected if any metric exceeds threshold.

\begin{table}[H]
  \centering
  \small
  \setlength{\tabcolsep}{8pt}
  \renewcommand{\arraystretch}{1.12}
  \begin{tabular}{lll}
    \toprule
    \textbf{Metric} & \textbf{Granularity} & \textbf{Max Threshold} \\
    \midrule
    Duplicate line fraction & Line & 0.30 \\
    Duplicate line character fraction & Line & 0.20 \\
    Duplicate paragraph fraction & Paragraph & 0.30 \\
    Duplicate paragraph character fraction & Paragraph & 0.20 \\
    Top 2-gram character fraction & Word n-gram & 0.20 \\
    Top 3-gram character fraction & Word n-gram & 0.18 \\
    Top 4-gram character fraction & Word n-gram & 0.16 \\
    Duplicate 5-gram character fraction & Word n-gram & 0.15 \\
    Duplicate 6-gram character fraction & Word n-gram & 0.14 \\
    Duplicate 7-gram character fraction & Word n-gram & 0.13 \\
    Duplicate 8-gram character fraction & Word n-gram & 0.12 \\
    Duplicate 9-gram character fraction & Word n-gram & 0.11 \\
    Duplicate 10-gram character fraction & Word n-gram & 0.10 \\
    \bottomrule
  \end{tabular}
  \caption{\textbf{Gopher Repetition Filter thresholds.}}
  \label{tab:d1_repetition_thresholds}
\end{table}

\paragraph{Stage 11: Custom Quality Filter}
FLUX introduces a novel document-level filter with three criteria applied conjunctively.

\begin{table}[H]
  \centering
  \small
  \setlength{\tabcolsep}{12pt}
  \renewcommand{\arraystretch}{1.15}
  \begin{tabular}{lcc}
    \toprule
    \textbf{Criterion} & \textbf{Bound} & \textbf{Threshold} \\
    \midrule
    Minimum token count & Min & 50 \\
    Stop-word ratio & Min & 0.20 \\
    Unclosed bracket ratio & Max & 0.05 \\
    \bottomrule
  \end{tabular}
  \caption{\textbf{Custom Quality Filter thresholds.}}
  \label{tab:d1_custom_thresholds}
\end{table}

\subsubsection{Stages 12--13: Line-level Quality Filter and Integrity Enforcement}
\paragraph{Stage 12: Line-level Quality Filter}
This stage removes lines matching boilerplate heuristics and returns cleaned documents. Documents reduced to zero lines are rejected.

\begin{table}[H]
  \centering
  \small
  \setlength{\tabcolsep}{6pt}
  \renewcommand{\arraystretch}{1.12}
  \begin{tabular}{p{3.4cm}p{10.0cm}}
    \toprule
    \textbf{Heuristic} & \textbf{Removal Criterion} \\
    \midrule
    Minimum word count & Line contains fewer than 2 whitespace-delimited words \\
    Maximum uppercase ratio & Uppercase characters exceed 50\% of total line characters \\
    Maximum numeric ratio & Numeric characters exceed 99.9999\% of total line characters \\
    Social engagement counters & Pattern: [0-9]+(K|M|B)? with likes, shares, comments, retweets, reposts, quotes, bookmarks, upvotes, downvotes, downloads, views, followers \\
    Substring modifiers & Line $\leq 10$ words and matches boilerplate markers such as ``items in cart'', ``Read more...'', ``Sign-in'' \\
    JS/CSS artifacts & Line-start code patterns (e.g., \texttt{function(}, \texttt{var}, \texttt{let}, \texttt{const}, \texttt{\$.}, \texttt{@media}, \texttt{=>}) \\
    Navigation/breadcrumb & Two or more segments split by \texttt{>}, \texttt{\guillemotright}, \texttt{/}, or \texttt{|} \\
    Cookie/GDPR banners & Contains cookie-consent/GDPR boilerplate phrases \\
    Social media CTAs & Line-start social prompts such as ``Follow us'', ``Subscribe now'', ``Share this'' \\
    Form element remnants & Full-line UI labels such as ``Username'', ``Password'', ``Email address'', ``Submit'', ``Register'' \\
    Timestamp/date-only lines & Line is only a date/time expression (DD/MM/YYYY, YYYY-MM-DD, HH:MM[:SS][AM|PM]) \\
    \bottomrule
  \end{tabular}
  \caption{\textbf{Line-level cleaning heuristics: removal criteria for all eleven heuristic classes.} Each class targets a distinct boilerplate pattern; lines matching any criterion are excised individually rather than triggering whole-document rejection.}
  \label{tab:d1_line_heuristics}
\end{table}

\paragraph{Stage 13: Word Removal Ratio Filter}
Post-cleaning integrity gate. Let $W_{\mathrm{pre}}$ and $W_{\mathrm{post}}$ denote word counts before and after Stage 12.
\[
\rho = \frac{W_{\mathrm{pre}}-W_{\mathrm{post}}}{W_{\mathrm{pre}}}
\]
A document is rejected if $\rho > 0.05$. Documents with $\rho \leq 0.05$ form the retained output.

\appsubsection{Per-Filter Document and Token Removal Statistics}
\label{app:filter_d2}
Table~\ref{tab:d2_per_filter_stats} reports document and token removal counts for every stage and sub-filter in FLUX over 10{,}000 WARC Files from CC-MAIN-2025-13 snapshot (GPT-2 tokenization).

\begin{table}[H]
  \centering
  \scriptsize
  \setlength{\tabcolsep}{5pt}
  \renewcommand{\arraystretch}{1.1}
  \begin{tabular}{p{3.0cm}p{1.8cm}p{1.8cm}p{3.0cm}p{1.4cm}}
    \toprule
    \textbf{Filter Stage} & \textbf{Type} & \textbf{Docs Removed} & \textbf{Tokens Removed} & \textbf{\% Total} \\
    \midrule
    \textbf{Initial Input} & --- & --- & 424{,}893{,}496{,}759 & 100.00\% \\
    UT1 Domain Blocklist (2) & Filter & 12{,}057 & 16{,}752{,}457 & 0.26\% \\
    URL Strict Substring (3) & Filter & 343 & 514{,}779 & 0.01\% \\
    URL Hard Substring (4) & Filter & 2{,}545 & 3{,}723{,}647 & 0.06\% \\
    URL Soft Substring (5) & Filter & 396 & 619{,}897 & 0.01\% \\
    URL Token Removal\textsuperscript{$\dagger$} (6) & Modifier & --- & 73{,}019{,}753 & 1.13\% \\
    Newline Normalization & Modifier & --- (modifier) & 657{,}924 & <0.01\% \\
    Language Identification & Filter & 156{,}394{,}512 & 343{,}974{,}627{,}759 & 80.96\% \\
    \textbf{Gopher Quality Filter} & \textbf{Group} & \textbf{12{,}636{,}392} & \textbf{5{,}801{,}555{,}757} & \textbf{1.37\%} \\
    \hspace{2mm}\textit{TooFewWords} & Sub-filter & 9{,}027{,}536 & 468{,}754{,}273 & 0.11\% \\
    \hspace{2mm}\textit{AlphaWordsRatio} & Sub-filter & 2{,}358{,}824 & 2{,}253{,}870{,}516 & 0.53\% \\
    \hspace{2mm}\textit{EllipsisLineRatio} & Sub-filter & 708{,}916 & 553{,}314{,}738 & 0.13\% \\
    \hspace{2mm}\textit{TooFewStopWords} & Sub-filter & 406{,}323 & 108{,}725{,}985 & 0.03\% \\
    \hspace{2mm}\textit{SymbolWordRatio} & Sub-filter & 69{,}650 & 87{,}859{,}905 & 0.02\% \\
    \hspace{2mm}\textit{BulletLineRatio} & Sub-filter & 13{,}077 & 43{,}425{,}129 & 0.01\% \\
    \hspace{2mm}\textit{AvgWordLen} & Sub-filter & 46{,}097 & 512{,}539{,}930 & 0.12\% \\
    \hspace{2mm}\textit{TooManyWords} & Sub-filter & 5{,}969 & 1{,}773{,}065{,}281 & 0.42\% \\
    \textbf{Nemo Filter} & \textbf{Group} & \textbf{191{,}825} & \textbf{425{,}774{,}107} & \textbf{0.10\%} \\
    \hspace{2mm}\textit{NonAlphaNumericRatio} & Sub-filter & 191{,}214 & 424{,}742{,}773 & 0.10\% \\
    \hspace{2mm}\textit{WhitespaceRatio} & Sub-filter & 288 & 444{,}061 & <0.01\% \\
    \hspace{2mm}\textit{ParenthesesRatio} & Sub-filter & 323 & 587{,}273 & <0.01\% \\
    \textbf{Gopher Repetition} & \textbf{Group} & \textbf{14{,}878{,}941} & \textbf{19{,}408{,}902{,}927} & \textbf{4.57\%} \\
    \hspace{2mm}\textit{DupLineFrac} & Sub-filter & 6{,}083{,}261 & 11{,}021{,}657{,}524 & 2.59\% \\
    \hspace{2mm}\textit{DupNGramCharFrac} & Sub-filter & 4{,}888{,}002 & 6{,}122{,}089{,}867 & 1.44\% \\
    \hspace{2mm}\textit{DupLineCharFrac} & Sub-filter & 1{,}986{,}819 & 1{,}511{,}334{,}597 & 0.36\% \\
    \hspace{2mm}\textit{TopNGramCharFrac} & Sub-filter & 1{,}919{,}888 & 752{,}283{,}657 & 0.18\% \\
    \hspace{2mm}\textit{DupParFrac} & Sub-filter & 605 & 1{,}531{,}837 & <0.01\% \\
    \textbf{Custom Quality Filter} & \textbf{Group} & \textbf{6{,}825{,}719} & \textbf{3{,}827{,}552{,}491} & \textbf{0.90\%} \\
    \hspace{2mm}\textit{StopWordRatio} & Sub-filter & 6{,}605{,}336 & 3{,}511{,}979{,}618 & 0.83\% \\
    \hspace{2mm}\textit{UnclosedBracketRatio} & Sub-filter & 220{,}383 & 315{,}572{,}873 & 0.07\% \\
    \textbf{Line Level Quality} & \textbf{Modifier} & \textbf{1{,}163} & \textbf{1{,}357{,}643{,}165} & \textbf{0.32\%} \\
    \hspace{2mm}\textit{UppercaseRatio} & Sub-filter & --- & 347{,}889{,}409 & 0.08\% \\
    \hspace{2mm}\textit{LineLength} & Sub-filter & --- & 579{,}523{,}752 & 0.14\% \\
    \hspace{2mm}\textit{CounterLine} & Sub-filter & --- & 4{,}898{,}313 & <0.01\% \\
    \hspace{2mm}\textit{Navigation} & Sub-filter & --- & 41{,}547{,}332 & 0.01\% \\
    \hspace{2mm}\textit{CookieBanner} & Sub-filter & --- & 75{,}172{,}871 & 0.02\% \\
    \hspace{2mm}\textit{SocialCTA} & Sub-filter & --- & 34{,}346{,}316 & 0.01\% \\
    \hspace{2mm}\textit{Timestamp} & Sub-filter & --- & 3{,}809{,}673 & <0.01\% \\
    Word Removal Ratio & Filter & 6{,}500{,}968 & 2{,}988{,}939{,}258 & 0.70\% \\
    \rowcolor{blue!10} \textbf{Final Retained Corpus (pre-dedup)} & \textbf{Output} & \textbf{37{,}947{,}106} & \textbf{37{,}563{,}276{,}357} & \textbf{8.84\%} \\
    \bottomrule
  \end{tabular}
  \caption{\textbf{Per-filter document and token removal statistics for FLUX over 10{,}000 WARC Files from CC-MAIN-2025-13 snapshot.}}
  \label{tab:d2_per_filter_stats}
\end{table}

\appsubsection{Pipeline Ablation: Four-Stage Design Evolution}
\label{app:filter_d3}
To characterize the contribution of each subsystem, we evaluate four configurations of increasing sophistication on 10{,}000 held-out WARC Files from CC-MAIN-2025-13 snapshot. Post-deduplication token yield and downstream aggregate score (530M scale; 10.6B training tokens) are summarized below.

\subsubsection{Configuration Definitions}
\begin{table}[H]
  \centering
  \small
  \setlength{\tabcolsep}{6pt}
  \renewcommand{\arraystretch}{1.15}
  \begin{tabular}{>{\raggedright\arraybackslash}p{3.6cm} >{\centering\arraybackslash}p{2.0cm} >{\centering\arraybackslash}p{1.9cm} >{\centering\arraybackslash}p{1.9cm} >{\centering\arraybackslash}p{1.9cm}}
    \toprule
    \textbf{Filter Component} & \textbf{URL+LID} & \textbf{DocFilter} & \textbf{LineClean} & \textbf{FLUX} \\
    \midrule
    URL Blocklist & Y & Y & Y & Y \\
    URL Substring Filtering & Y & Y & Y & Y \\
    URL Token Removal & Y & Y & Y & Y \\
    Newline Normalization & Y & Y & Y & Y \\
    Language Identification & Y & Y & Y & Y \\
    Gopher Repetition & --- & Y & Y & Y \\
    Gopher Quality & --- & Y & Y & Y \\
    Nemo Filter & --- & Y & Y & Y \\
    Badwords Filter & --- & Y & Y & --- \\
    Line-level Cleaning (core) & --- & --- & Y & Y \\
    Line-level Cleaning (+6 heuristics) & --- & --- & --- & Y \\
    Custom Quality Filter & --- & --- & Y & Y \\
    Word Removal Ratio Filter & --- & --- & Y & Y \\
    \bottomrule
  \end{tabular}
  \caption{\textbf{Filter composition across pipeline configurations.} Y = active; --- = not applied.}
  \label{tab:d3_filter_composition}
\end{table}

\subsubsection{Stage-by-Stage Results}
\label{app:ablation_stages}

\begin{figure}[H]
  \centering
  \includegraphics[width=0.92\linewidth]{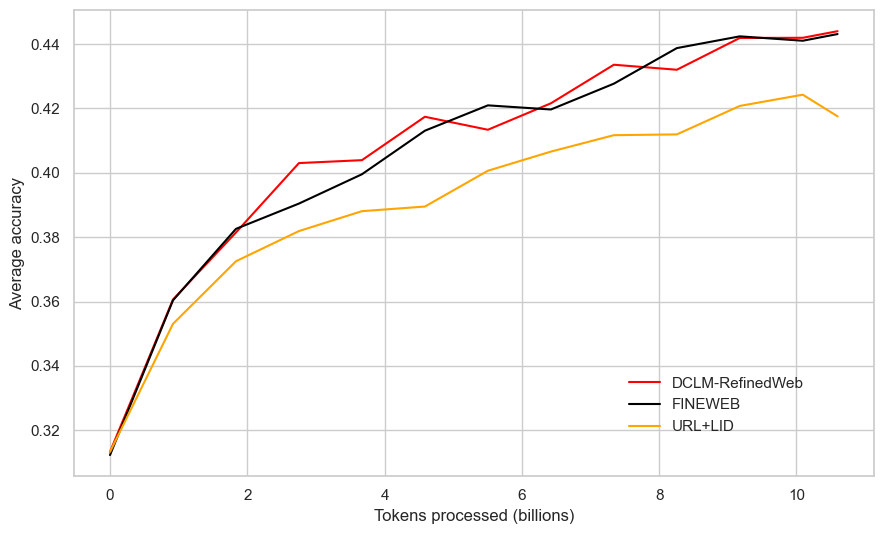}
  \caption{\textbf{Aggregate learning curves for URL+LID against DCLM-RefinedWeb and FineWeb (530M scale; 10.6B tokens).}}
  \label{fig:d3_stage1_placeholder}
\end{figure}

\paragraph{Stage 1 --- URL+LID: volume without quality control is insufficient.}
URL+LID retains the largest token volume of all configurations: 42.4B post-deduplication tokens,
representing a 70.9\% surplus over DCLM-RefinedWeb and 73.0\% over FineWeb. Despite this
retention advantage, the downstream aggregate score of 41.76 falls 2.64~pp below DCLM-RefinedWeb
and 2.55~pp below FineWeb. As shown in Figure~\ref{fig:d3_stage1_placeholder}, URL+LID trails both baselines
across the full training budget. This result demonstrates that raw token volume without quality
filtering does not translate into downstream capability: the presence of noise and low-information
content actively suppresses training efficiency regardless of corpus scale.

\begin{figure}[H]
  \centering
  \includegraphics[width=0.92\linewidth]{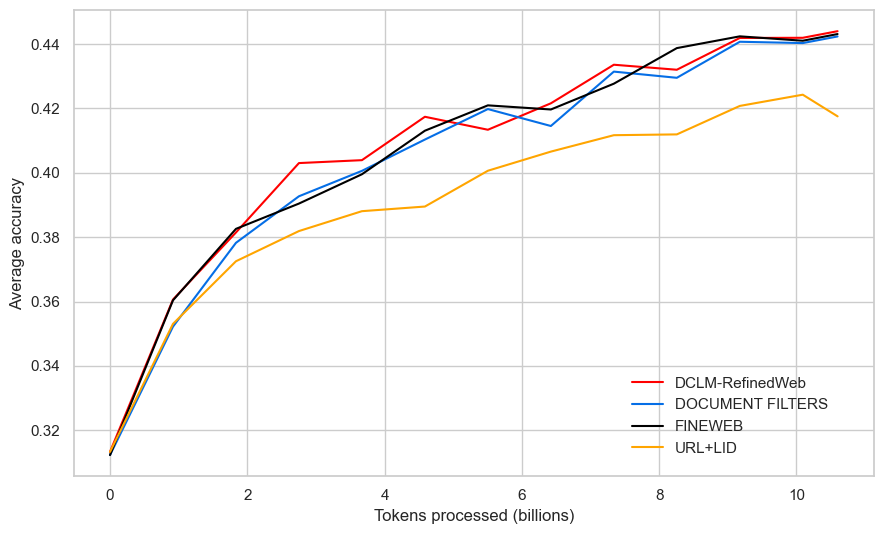}
  \caption{\textbf{Aggregate learning curves for DocFilter against DCLM-RefinedWeb and FineWeb (530M scale; 10.6B tokens).} }
  \label{fig:d3_stage2_placeholder}
\end{figure}

\paragraph{Stage 2 --- DocFilter: quality parity at severe retention cost.}
Augmenting URL+LID with the Gopher, Nemo, and C4 Badwords filters raises the aggregate score to
44.24, closing to within 0.16~pp of DCLM-RefinedWeb (44.40) and 0.07~pp of FineWeb (44.31). As
shown in Figure~\ref{fig:d3_stage2_placeholder}, DocFilter tracks closely alongside both baselines throughout
training. However, this quality recovery comes at a steep cost to retention: post-deduplication
tokens collapse to 23.4B, a 44.8\% reduction from URL+LID and 5.7\% below DCLM-RefinedWeb. The
document-level rejection paradigm can match external baselines but cannot exceed them: by
discarding entire documents in which only a fraction of lines are noisy, it permanently loses the
high-quality content that coexists with the noise.

\begin{figure}[H]
  \centering
  \includegraphics[width=0.92\linewidth]{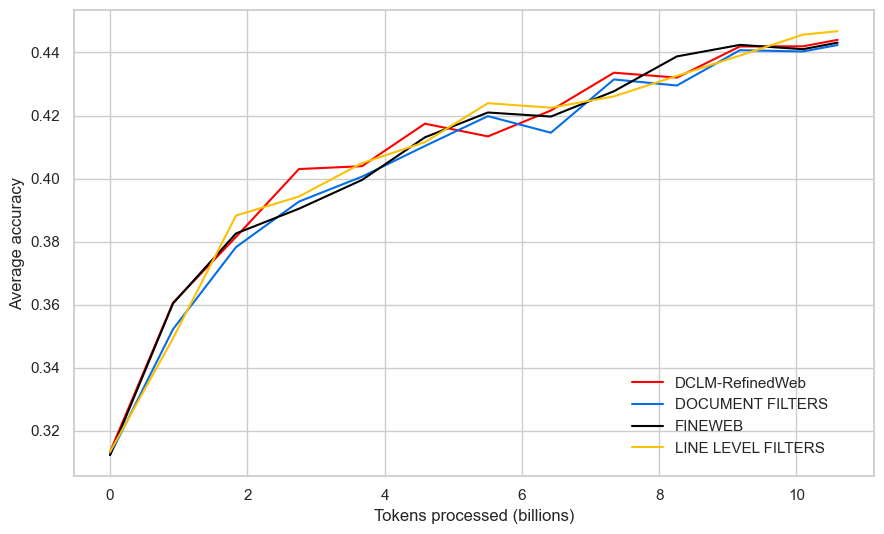}
  \caption{\textbf{Aggregate learning curves for LineClean against DCLM-RefinedWeb and FineWeb (530M scale; 10.6B tokens).}}
  \label{fig:d3_stage3_placeholder}
\end{figure}

\paragraph{Stage 3 --- LineClean: first configuration to surpass both baselines.}
Replacing document-level C4 Badwords rejection with targeted line-level excision, and introducing
the Custom Quality Filter, produces the first configuration to outperform both external baselines.
LineClean achieves an aggregate score of 44.68, exceeding DCLM-RefinedWeb by 0.28~pp and FineWeb
by 0.37~pp, as shown in Figure~\ref{fig:d3_stage3_placeholder}. A retention penalty persists, however:
post-deduplication tokens stand at 21.0B, 15.4\% below DCLM-RefinedWeb, because line-level
cleaning is stacked on top of the remaining document-level filters in this configuration,
compounding the rejection penalty.

\begin{figure}[H]
  \centering
  \includegraphics[width=0.92\linewidth]{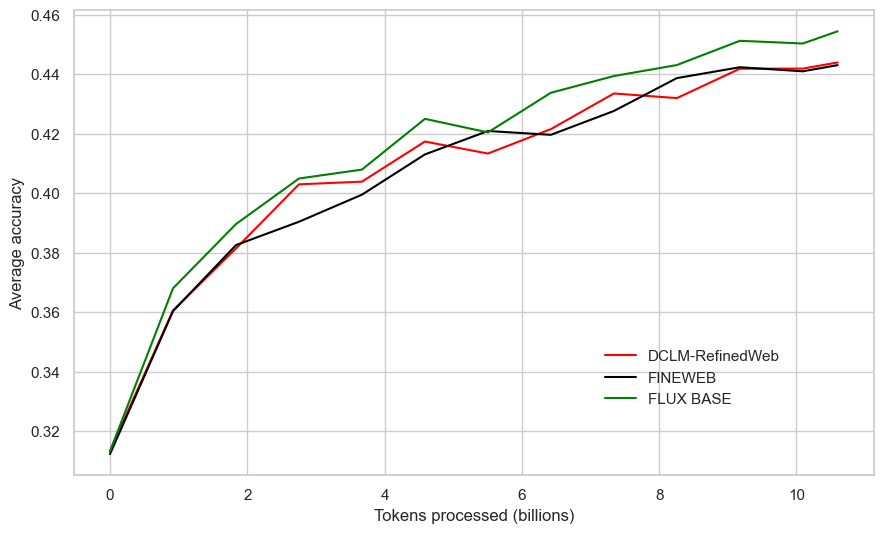}
  \caption{\textbf{Aggregate learning curves for FLUX-Base against DCLM-RefinedWeb and FineWeb (530M scale; 10.6B tokens).}}
  \label{fig:d3_stage4_placeholder}
\end{figure}

\paragraph{Stage 4 --- FLUX: highest quality and highest retention simultaneously.}
Removing C4 Badwords from the document level entirely and extending line-level cleaning with six
additional targeted heuristics resolves the retention penalty of LineClean while further improving
downstream quality. FLUX achieves an aggregate score of 45.45 --- the highest of all four
configurations --- while recovering 27.2B post-deduplication tokens, 16.4\% more than DocFilter
and 9.7\% more than DCLM-RefinedWeb. As shown in Figure~\ref{fig:d3_stage4_placeholder}, FLUX leads both
baselines across the full training budget, with the margin widening steadily beyond 6B tokens.
FLUX is the only configuration in this progression that strictly dominates both external baselines
on both token retention and downstream quality simultaneously (Table~\ref{tab:d3_consolidated_results}).

The key mechanism is the removal of C4 Badwords from the document level: enforcing those noise
classes at line resolution recovers documents that would otherwise have been discarded entirely,
while the six extended heuristics surgically excise the specific lines that made those documents
problematic.

\subsubsection{Consolidated Results}
\begin{table}[H]
  \centering
  \small
  \setlength{\tabcolsep}{6pt}
  \renewcommand{\arraystretch}{1.15}
  \resizebox{\linewidth}{!}{%
  \begin{tabular}{lcccccc}
    \toprule
    \textbf{Configuration} & \textbf{Tokens} & \makecell{\textbf{vs DCLM-}\\\textbf{RefinedWeb}} & \textbf{vs FineWeb} & \textbf{Aggregate} & \makecell{\textbf{vs DCLM-}\\\textbf{RefinedWeb}} & \textbf{vs FineWeb} \\
    \midrule
    URL+LID & 42.39B & \textcolor{green!60!black}{+70.9\%} & \textcolor{green!60!black}{+73.0\%} & 41.76 & \textcolor{red!70!black}{-2.64 pp} & \textcolor{red!70!black}{-2.55 pp} \\
    DocFilter & 23.39B & \textcolor{red!70!black}{-5.7\%} & \textcolor{red!70!black}{-4.5\%} & 44.24 & \textcolor{red!70!black}{-0.16 pp} & \textcolor{red!70!black}{-0.07 pp} \\
    LineClean & 21.00B & \textcolor{red!70!black}{-15.4\%} & \textcolor{red!70!black}{-14.3\%} & 44.68 & \textcolor{green!60!black}{+0.28 pp} & \textcolor{green!60!black}{+0.37 pp} \\
    \rowcolor{blue!10} FLUX-Base & 27.22B & \textcolor{green!60!black}{+9.7\%} & \textcolor{green!60!black}{+11.1\%} & 45.45 & \textcolor{green!60!black}{+1.05 pp} & \textcolor{green!60!black}{+1.14 pp} \\
    FineWeb & 24.50B & --- & --- & 44.31 & --- & --- \\
    DCLM-RefinedWeb & 24.81B & --- & --- & 44.40 & --- & --- \\
    \bottomrule
  \end{tabular}
  }
  \caption{\textbf{Post-deduplication token retention and downstream aggregate score for all four pipeline configurations with DCLM and FineWeb reference points.} FLUX is the only configuration that exceeds both DCLM-RefinedWeb and FineWeb on token retention and downstream quality simultaneously; all other configurations sacrifice one for the other.}
  \label{tab:d3_consolidated_results}
\end{table}

\newpage
\appsubsection{Language Identification Strategy Ablation}
\label{app:lid_ablation}
We compare two LID scoring strategies and evaluate impact on corpus size and downstream model quality.

\subsubsection{Formulations}
Whole-document LID removes all newlines and scores the full text with FastText \texttt{lid.176.bin}; retain iff predicted language is English and confidence meets threshold $\tau=0.65$.

Weighted average line LID scores each line independently and computes:
\[
\text{score}_{\mathrm{en}} = \frac{\sum_i |l_i|\, p_{\mathrm{en}}(l_i)}{\sum_i |l_i|}
\]
where $|l_i|$ is byte length of line $i$.

\subsubsection{Why the Two Strategies Differ}
\begin{table}[H]
  \centering
  \small
  \setlength{\tabcolsep}{6pt}
  \renewcommand{\arraystretch}{1.15}
  \begin{tabular}{>{\raggedright\arraybackslash}p{5.2cm} >{\centering\arraybackslash}p{1.8cm} >{\centering\arraybackslash}p{1.8cm} >{\raggedright\arraybackslash}p{3.4cm}}
    \toprule
    \textbf{Document Composition} & \textbf{Whole-Doc Score} & \textbf{Weighted Score} & \textbf{Retention Decision} \\
    \midrule
    800 B English ($p=0.95$) + 200 B French ($p=0.10$) & 0.88 & 0.78 & Both retain \\
    400 B English ($p=0.80$) + 600 B French ($p=0.10$) & 0.70 & 0.38 & Whole retains; Weighted rejects \\
    \bottomrule
  \end{tabular}
  \caption{\textbf{LID scores for two mixed-language documents under each strategy.} Weighted average line LID rejects a majority-French document that whole-document LID retains, illustrating how line-level scoring can discard documents that still carry useful English-language modeling signal.}
  \label{tab:d4_mixed_language_examples}
\end{table}

\subsubsection{Retention Statistics}
\begin{table}[H]
  \centering
  \small
  \setlength{\tabcolsep}{5pt}
  \renewcommand{\arraystretch}{1.15}
  \begin{tabular}{>{\raggedright\arraybackslash}p{2.8cm} >{\centering\arraybackslash}p{2.2cm} >{\centering\arraybackslash}p{2.0cm} >{\centering\arraybackslash}p{2.0cm} >{\centering\arraybackslash}p{2.3cm}}
    \toprule
    \textbf{LID Strategy} & \textbf{Documents Retained} & \textbf{Tokens Retained} & \textbf{Avg. Docs/JSONL} & \textbf{Avg. Tokens/JSONL} \\
    \midrule
    Whole-Document & 49.4M & 48.1B & 5{,}512 & 5{,}360{,}516 \\
    Weighted Avg. Line & 47.0M & 45.0B & 5{,}239 & 5{,}088{,}583 \\
    \bottomrule
  \end{tabular}
  \caption{\textbf{Retention statistics for each LID strategy across 10,000 WARC Files from CC-MAIN-2025-13 snapshot.} Whole-document LID retains 2.4M more documents and 3.1B more tokens than weighted average line LID, motivating its adoption as the FLUX default.}
  \label{tab:d4_lid_retention}
\end{table}

\subsubsection{Downstream Model Quality}
\begin{figure}[H]
  \centering
  \includegraphics[width=0.92\linewidth]{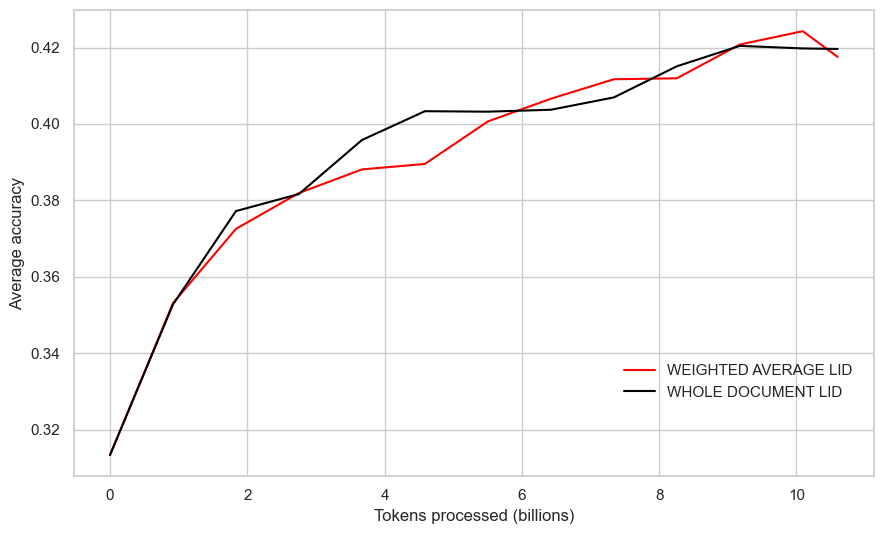}
  \caption{\textbf{Whole-document LID (black) versus Weighted Average line LID (red) (530M scale; 10.6B tokens).}}
  \label{fig:d4_lid_quality_placeholder}
\end{figure}

\begin{table}[H]
  \centering
  \small
  \setlength{\tabcolsep}{14pt}
  \renewcommand{\arraystretch}{1.15}
  \begin{tabular}{lc}
    \toprule
    \textbf{LID Strategy} & \textbf{Aggregate} \\
    \midrule
    Whole-Document LID & 0.4191 \\
    Weighted Average Line LID & 0.4136 \\
    \bottomrule
  \end{tabular}
  \caption{\textbf{Aggregate accuracy (530M scale; 10.6B tokens).} Whole-document LID outperforms weighted average line LID on aggregate score, confirming that its retention advantage translates to better downstream model quality.}
  \label{tab:d4_lid_downstream}
\end{table}

\subsubsection{Discussion}
Whole-document LID retains more data and produces better models. Weighted line LID is stricter but can remove documents containing short non-English spans that still carry useful English-language modeling signal. We therefore adopt whole-document LID as the FLUX default.

\newpage

\appsubsection{Computational Cost}
Preprocessing cost is a practical constraint at web scale. We benchmark FLUX, DCLM, and FineWeb filtering on identical hardware (\texttt{c8a.8xlarge}, 32 vCPUs) over 10{,}000 WARC Files from CC-MAIN-2025-13 snapshot~\citep{li_datacomp-lm_2025,penedo_fineweb_2024}.

\begin{table}[H]
  \centering
  \small
  \setlength{\tabcolsep}{8pt}
  \renewcommand{\arraystretch}{1.15}
  \begin{tabular}{lccccc}
    \toprule
    \textbf{Pipeline} & \textbf{Input} & \textbf{Wall-clock} & \textbf{CPU-hrs} & \textbf{CPU-hrs/TB} & \textbf{vs. FLUX} \\
    \midrule
    DCLM & 1.20 TB & 11 hr 7 min & 355.7 & 296.4 & 6.26$\times$ \\
    FineWeb & 0.60 TB & 5 hr 39 min & 181.0 & 301.3 & 6.36$\times$ \\
    \rowcolor{blue!10} FLUX & 1.03 TB & 1 hr 31 min & 48.5 & 47.3 & 1.00$\times$ \\
    \bottomrule
  \end{tabular}
  \caption{\textbf{Filtering-stage compute cost (c8a.8xlarge, 32 vCPUs, 10,000 WARC Files from CC-MAIN-2025-13 snapshot).} FLUX is 6.26$\times$ more efficient than DCLM and 6.36$\times$ more efficient than FineWeb on a per-TB basis, completing filtering of 1.03 TB in under 2 hours.}
  \label{tab:d5_filter_compute_cost}
\end{table}

\appsection{Bloom Filter Deduplication: Configuration and Scale}
\label{app:dedup}

\appsubsection{Overview}
\label{sec:dedup_details}
We apply Bloom Filter-based Fuzzy Deduplication (BFF) to remove redundant content at both the paragraph and document level~\citep{bloom_spacetime_1970,broder1997resemblance,lee_deduplicating_2022}. The pipeline is implemented in Rust and operates on JSONL-formatted documents. Each document is split into paragraphs, and every paragraph is tokenized into overlapping n-gram shingles of a fixed size. Each shingle is hashed and checked against a shared Bloom filter. If more than 80\% of a paragraph's shingles are already present in the filter, the paragraph is treated as a near-duplicate and discarded. Only novel paragraphs have their n-grams inserted back into the filter, ensuring removed content does not pollute the seen-set. At the document level, if enough paragraphs within a single document are flagged as duplicates, the entire document is dropped rather than partially pruned. This combined \texttt{OldBoth} mode---paragraph removal with a document-level fallback---balances precision and recall in duplicate removal.

Two key parameters govern the Bloom filter's behavior: the \textbf{false positive (FP) rate}, which controls memory consumption and duplicate aggressiveness, and the \textbf{expected n-gram count}, which sizes the filter to the anticipated corpus. Together these determine \textbf{BFF sparsity}---the fraction of bits set in the filter---which serves as our primary diagnostic for filter health. A sparsity near 0.5 is theoretically optimal: too low indicates under-utilization, too high indicates excessive collisions and an effective FP rate above target. We run ablations over both parameters to identify a stable operating point before deduplicating the full corpus.

\appsubsection{Corpus Scale and Cross-Dump Deduplication}
Our input consists of 1.4 TB from the CC-MAIN-2025-47 filtered dump (312{,}017{,}023{,}360 tokens) and 1.5 TB from the CC-MAIN-2025-51 filtered dump (289{,}852{,}248{,}915 tokens), totaling approximately 2.9 TB and \textasciitilde601.9B tokens. Deduplication is applied globally across both dumps by maintaining a single shared Bloom filter (\texttt{total\_shards = 1}), ensuring that duplicates appearing across dump boundaries are removed. The final deduplicated output is 1{,}795 GB, containing approximately 398B tokens---a retention rate of \textasciitilde66\% by data size. The remaining \textasciitilde34\% of content is identified as near-duplicate and discarded. This single-shard approach is conservative and appropriate for two dumps; extending to all 116 Common Crawl snapshots would require a multi-shard strategy because the aggregate n-gram volume would exceed available RAM for a single filter.

\appsubsection{False Positive Rate Ablation}
The FP rate is the probability that a novel n-gram is incorrectly identified as a duplicate. A lower FP rate requires a larger Bloom filter (more bits per n-gram) and more hash function evaluations per lookup, increasing both memory consumption and computation time. We ran ablations across six FP rate settings---from an extremely conservative $10^{-13}$ down to a highly permissive $10^{-1}$---to characterize the trade-off between deduplication fidelity and wall-clock runtime.

The results in Table~\ref{tab:dedup_fp_ablation} reveal a clear pattern. At the conservative end (FP = $10^{-13}$, 30B n-grams), the pipeline retains 944 GB but takes 2 hours and 15 minutes to run. Relaxing the FP rate to $10^{-6}$ and $10^{-3}$ (with appropriately sized n-gram counts) maintains the same 944 GB retention while cutting runtime to 54 minutes and 42 minutes respectively---a significant speedup with no loss in data quality. This is a key finding: \textbf{the FP rate can be relaxed substantially without impacting the amount or quality of retained data}, as long as the expected n-gram count is set appropriately.

However, pushing to FP = $10^{-1}$ with a low expected n-gram count (30B) causes retention to drop to only 865 GB---nearly 80 GB less than other configurations. This indicates the filter has become too coarse: with too few bits per n-gram, the false positive rate exceeds its nominal target, and unique content starts being incorrectly removed. The takeaway is that \textbf{there exists a practical lower bound on the FP rate below which data loss begins to occur}, and operating near this boundary is risky.

\appsubsection{Expected N-gram Count and BFF Sparsity}
While the FP rate controls the filter's theoretical precision, the expected n-gram count controls how the filter is sized relative to the actual data. If the true number of n-grams in the corpus exceeds the expected count, the filter overfills---bits become saturated, sparsity climbs well above 0.5, and the effective false positive rate rises far beyond its nominal target. This is precisely what we observe in configurations such as FP = $10^{-3}$ with only 30B expected n-grams: the sparsity reaches 0.911, meaning over 90\% of bits are set, which makes the filter nearly uninformative and causes it to remove more content than intended (936 GB retained vs. 944 GB for better-calibrated settings).

Conversely, setting the expected n-gram count too high wastes memory without improving filter quality, as the filter is pre-allocated for more n-grams than will ever arrive. The optimal configuration is one where the sparsity lands as close to 0.5 as possible after the full corpus has been processed.

Our ablations show that \textbf{FP = $10^{-3}$ with 100B expected n-grams} achieves a post-run sparsity of \textbf{0.516}---the closest to 0.5 across all tested configurations---while retaining 944 GB and completing in just \textbf{37 minutes}. This is our selected configuration. It represents the ideal operating point: the filter is correctly sized for the corpus, the FP rate is comfortably within a safe range, and the runtime is practical. For reference, FP = $10^{-2}$ with 100B n-grams achieves a faster 23-minute runtime but yields a sparsity of 0.578, indicating a slightly overfull filter. FP = $10^{-1}$ with 70B n-grams runs in 30 minutes but has sparsity of 0.708---too high to be reliable.

\begin{table}[H]
  \centering
  \small
  \setlength{\tabcolsep}{6pt}
  \renewcommand{\arraystretch}{1.15}
  \begin{tabular}{l l r l r r}
    \toprule
    \textbf{FP Rate} & \textbf{Expected n-grams} & \textbf{Retention (GB)} & \textbf{Time} & \textbf{Bin Size} & \textbf{Sparsity} \\
    \midrule
    $10^{-13}$ & 30B  & 944 & 2 hr 15 min & 216 & 0.914 \\
    $10^{-13}$ & 50B  & 944 & 1 hr 58 min & 362 & 0.771 \\
    $10^{-6}$  & 70B  & 944 & 54 min     & 234 & 0.716 \\
    $10^{-3}$  & 30B  & 936 & 42 min     & 50  & 0.911 \\
   \rowcolor{blue!10} $\mathbf{10^{-3}}$ & \textbf{100B} & \textbf{944} & \textbf{37 min} & \textbf{179} & \textbf{0.516} \\
    $10^{-2}$  & 100B & 944 & 23 min     & 56  & 0.578 \\
    $10^{-1}$  & 70B  & 939 & 30 min     & 39  & 0.708 \\
    \bottomrule
  \end{tabular}
  \caption{\textbf{Ablation results over FP rate and expected n-gram count.} The selected configuration (FP = $10^{-3}$, 100B expected n-grams) achieves a post-run sparsity of 0.516 --- closest to the theoretically optimal 0.5 --- while retaining 944 GB and completing in 37 minutes.}
  \label{tab:dedup_fp_ablation}
\end{table}

\appsection{Dual-Bin FastText Classification: Threshold Sweep and Operating Point Selection}
\label{sec:classifier_ablations}

For each threshold configuration, retention statistics are computed on a fixed input of 100B post-deduplication tokens from CC-MAIN-2025-51 (100,000,766,736 GPT-2 tokens across 51,103 files and 101,840,883 documents). The logical-OR acceptance rule is applied independently for each $(\tau_{\mathrm{DCLM}}, \tau_{\mathrm{BETR}})$ pair, and output token counts and document counts are recorded directly from the classifier output. Token retention percentage is computed as output tokens divided by total input tokens. Table~\ref{tab:threshold_sweep} reports the full sweep.

\begin{table}[H]
  \centering
  \small
  \setlength{\tabcolsep}{8pt}
  \renewcommand{\arraystretch}{1.15}
  \begin{tabular}{c c c c c}
    \toprule
    $\tau_{\mathrm{DCLM}}$ & $\tau_{\mathrm{BETR}}$ & \textbf{Retention \%} & \textbf{MMLU} & \textbf{Aggregate} \\
    \midrule
    0.2000 & 0.6350 & 26.43\% & 0.2887 & 0.5028 \\
  \rowcolor{blue!10}  \textbf{0.025119} & \textbf{0.7600} & \textbf{26.88\%} & \textbf{0.2982} & \textbf{0.5065} \\
    0.0900 & 0.6350 & 29.25\% & 0.2920 & 0.5022 \\
    0.018112 & 0.7347 & 31.10\% & 0.2949 & 0.5025 \\
    0.018112 & 0.7000 & 33.32\% & 0.2915 & 0.5012 \\
    \bottomrule
  \end{tabular}
  \caption{\textbf{Full Dual-Bin FastText classification threshold sweep (100B tokens post-dedup input).} The selected operating point $(\tau_{\mathrm{DCLM}} = 0.025119, \tau_{\mathrm{BETR}} = 0.76)$ achieves the highest Aggregate and MMLU scores across all evaluated pairs; configurations retaining more tokens yield strictly lower scores on both metrics.}
  \label{tab:threshold_sweep}
\end{table}

The selected configuration (bold) achieves the highest Aggregate and MMLU while retaining only 26.88\% of input tokens. Configurations retaining more tokens (31.10\%, 33.32\%) yield strictly lower scores on both metrics, confirming that quality and retention are not monotonically aligned and that threshold selection were grounded in downstream evaluation rather than retention targets.

\appsection{Corpus Composition and Domain Analysis}
\label{app:corpus_composition}
\label{sec:dataset_analysis}

\begin{figure}[H]
  \centering
  \includegraphics[width=\textwidth,trim=0 0 0 55,clip]{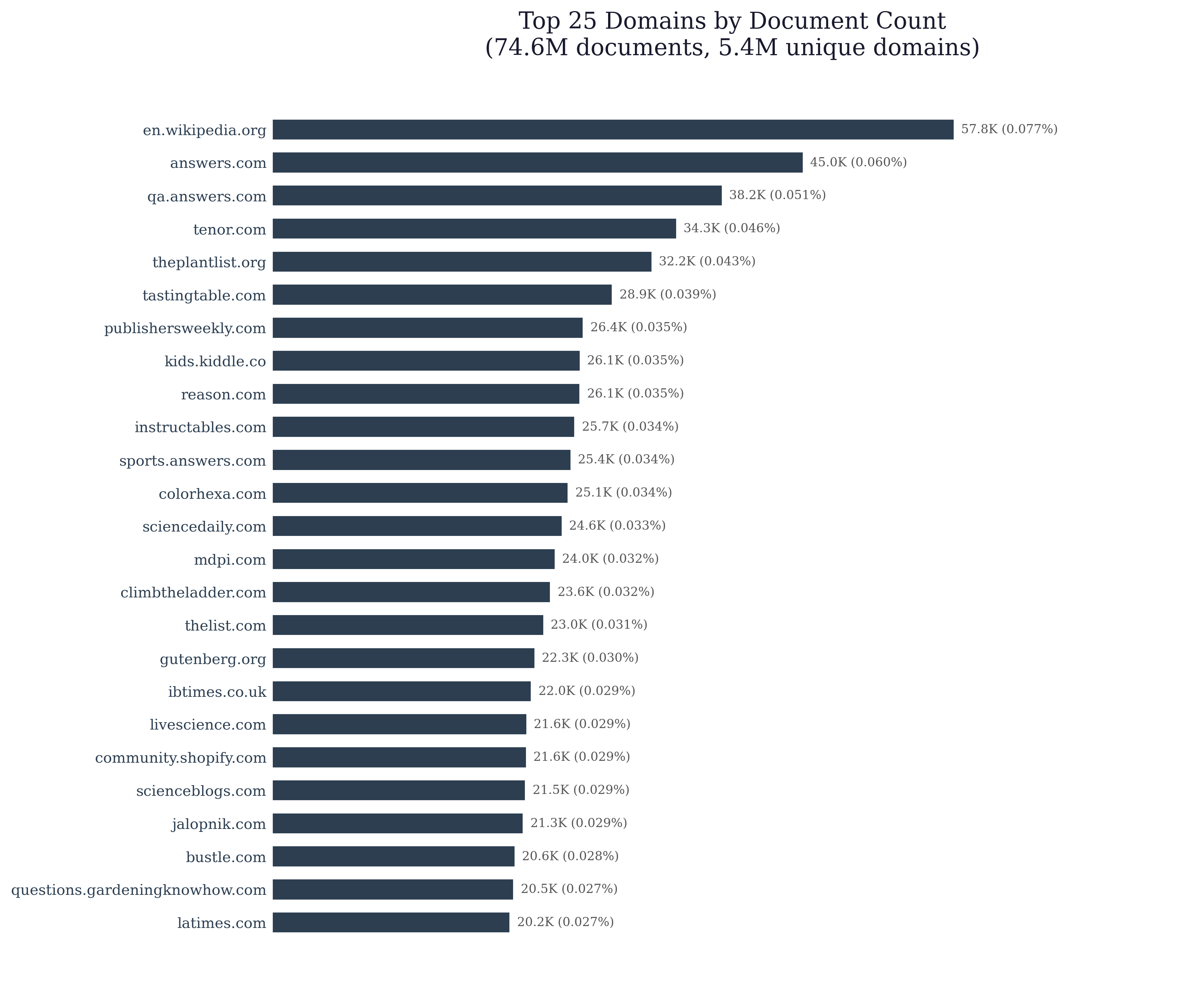}
  \caption{\textbf{Overview of corpus composition in the FLUX dataset.} The figure provides a high-level characterization of the FLUX corpus, showing how documents are distributed across major topical domains and structural formats. The diversity observed reflects the broad coverage of web content captured through the FLUX preprocessing pipeline, with detailed domain and format analyses reported in Figures~17 and~18.}
  \label{fig:top_domains_distribution}
\end{figure}

\begin{figure}[H]
  \centering
  \includegraphics[width=\textwidth]{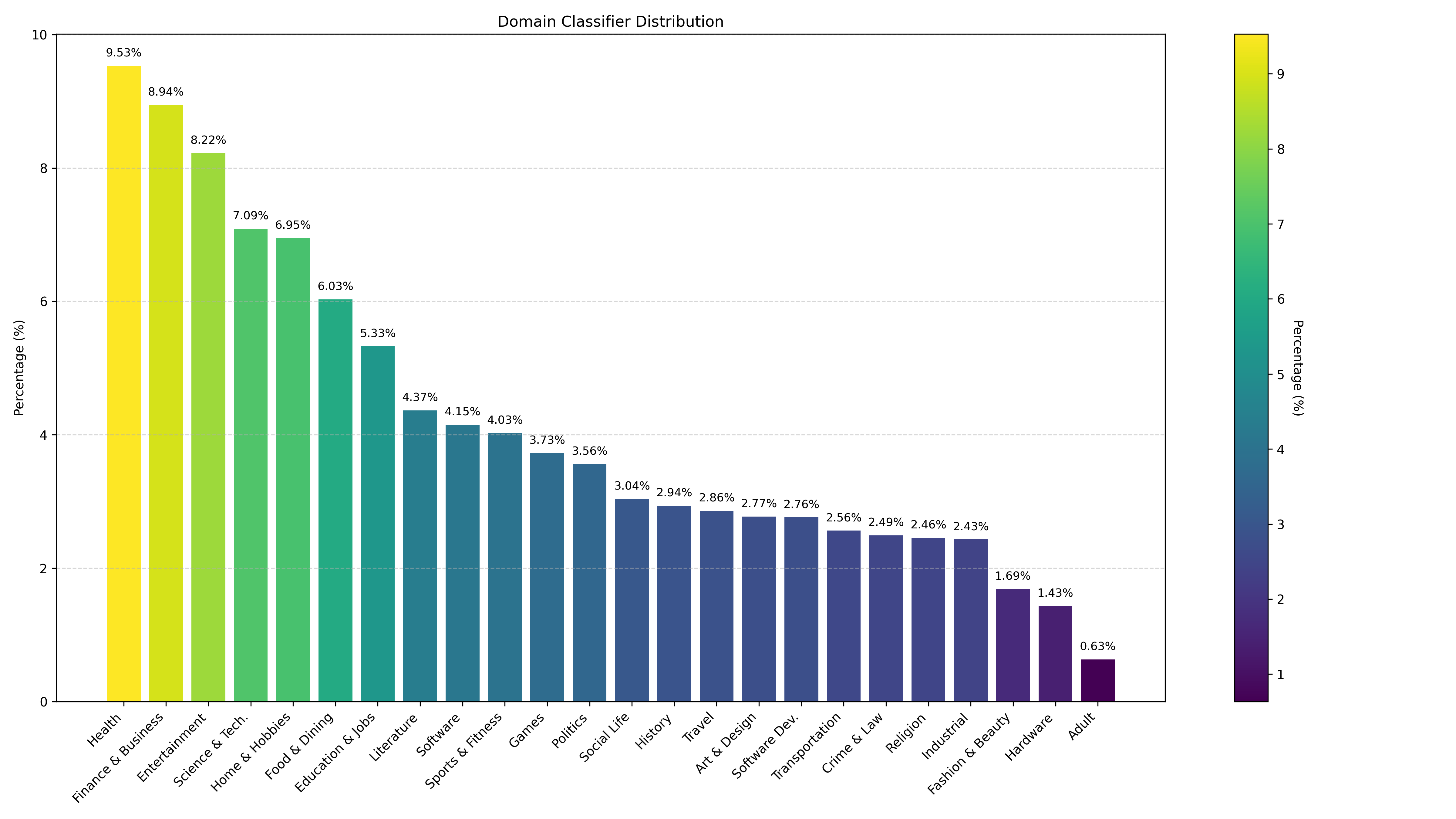}
  \caption{\textbf{Domain category distribution.} The FLUX corpus exhibits a diverse domain composition spanning informational, technical, and lifestyle-oriented categories. Health (9.53\%), Finance \& Business (8.94\%), Entertainment (8.22\%), and Science \& Tech (7.09\%) represent the largest segments. Additional domains such as Home \& Hobbies, Food \& Dining, and Education \& Jobs contribute substantial coverage, while categories including Politics, Social Life, and History appear at moderate frequencies. Sensitive domains remain minimal (Adult: 0.63\%), indicating effective suppression of undesirable content during dataset construction.}
  \label{fig:web_organizer_domain_distribution}
\end{figure}

\begin{figure}[H]
  \centering
  \includegraphics[width=\textwidth]{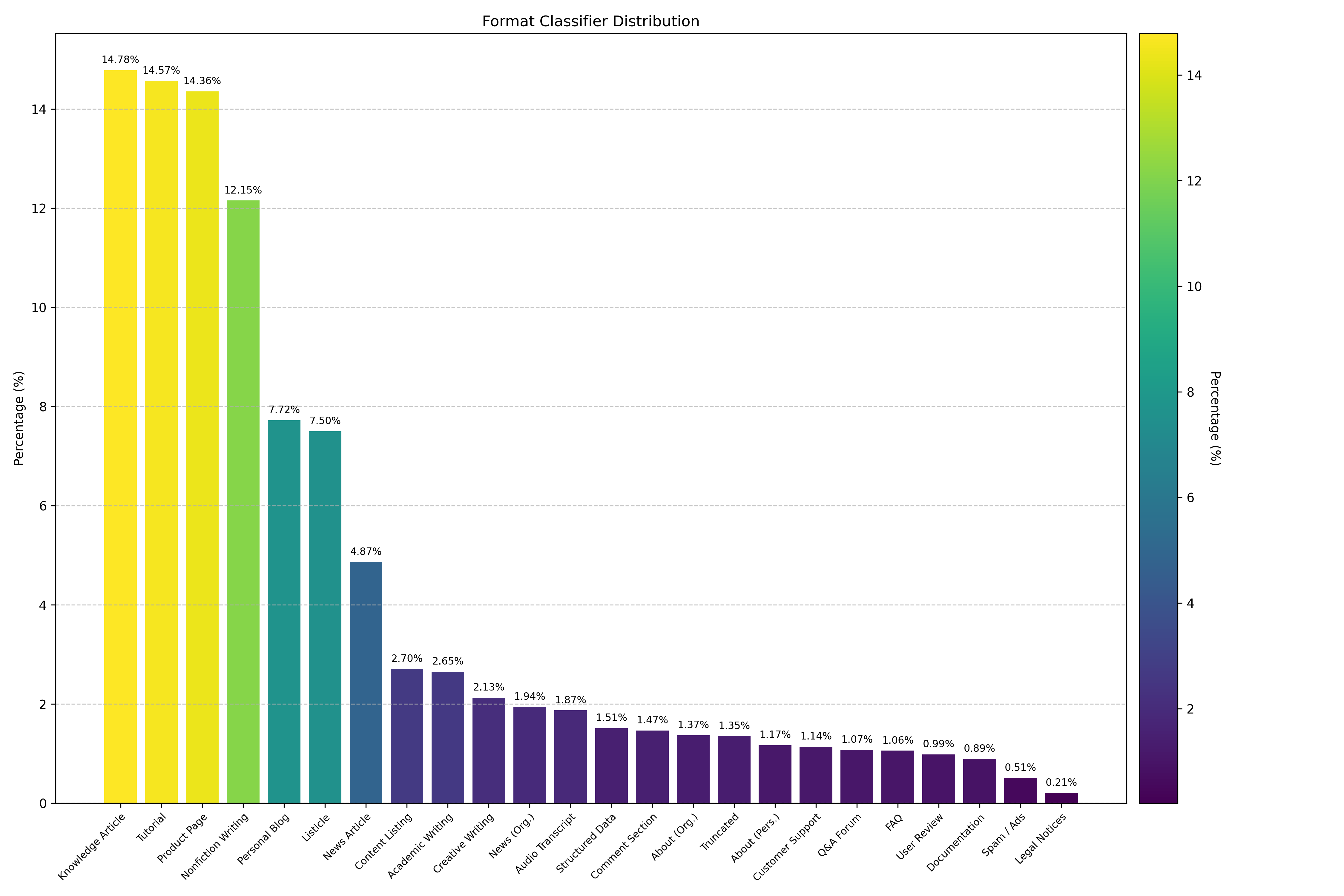}
  \caption{\textbf{Format category distribution.} The FLUX dataset contains a diverse mix of web content formats. Informational structures dominate the corpus, with Knowledge Articles (14.78\%), Tutorials (14.57\%), and Product Pages (14.36\%) forming the largest shares. Narrative and blog-style formats, including Nonfiction Writing (12.15\%) and Personal Blogs (7.72\%), also contribute significantly. Additional formats such as Listicles (7.50\%) and News Articles (4.87\%) further enrich structural diversity. Low-value categories including Spam/Ads (0.51\%) and Legal Notices (0.21\%) remain minimally represented, reflecting the effectiveness of the filtering pipeline.}
  \label{fig:web_organizer_format_distribution}
\end{figure}

To better understand the characteristics of the FLUX dataset, we analyze its corpus composition across topical domains and structural content formats. Following the domain- and format-based organization framework proposed by \citet{wettig_organize_2025}, we categorize documents using lightweight domain and format classifiers. Figure~\ref{fig:top_domains_distribution} provides an overview of the most frequent domains observed in the corpus. Figures~\ref{fig:web_organizer_domain_distribution} and \ref{fig:web_organizer_format_distribution} further break down the dataset by domain and format categories, illustrating the broad topical and structural diversity of the FLUX corpus while confirming that low-value content categories remain minimal after preprocessing.

\clearpage
\appsection{Extended Evaluation Results}
\label{app:extended_eval}
\label{sec:extended_eval}

We report per-benchmark learning curves at 3B scale and full evaluation results at the 530M and 1B scales. These results are consistent with the 3B-scale findings reported in Section~\ref{sec:experiments} and confirm that dataset quality rankings are stable across compute scales.

\begin{figure}[H]
  \centering
  \includegraphics[width=\linewidth]{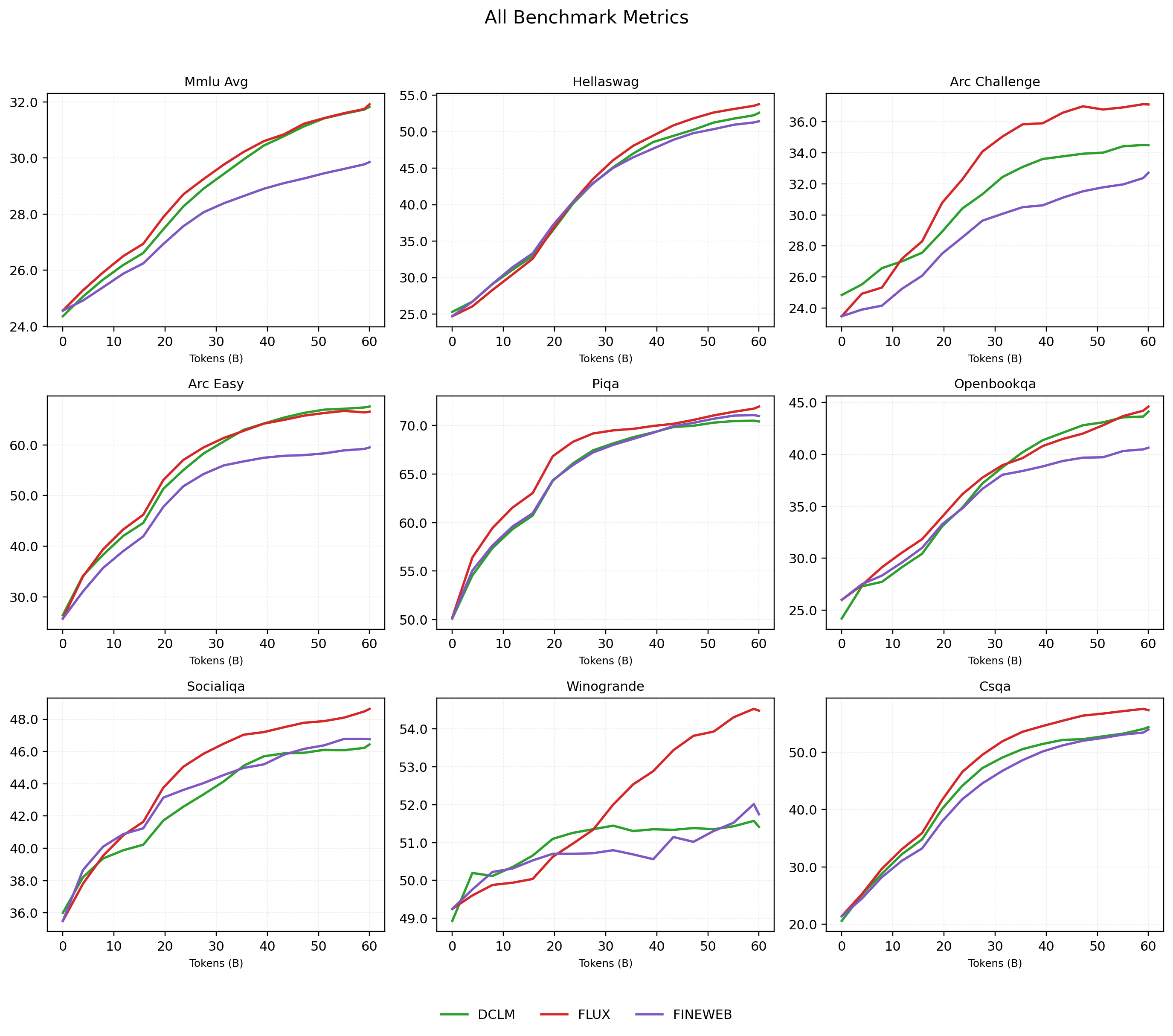}
  \caption{\textbf{Learning curves for FLUX, DCLM, and FineWeb across all nine evaluation benchmarks (3B scale; 60B tokens).}}
  \label{fig:olmes_9_metrics}
\end{figure}

\begin{table}[H]
  \centering
  \footnotesize
  \setlength{\tabcolsep}{4pt}
  \renewcommand{\arraystretch}{1.15}
  \begin{tabular}{l S[table-format=1.3] S[table-format=1.3] S[table-format=1.3] S[table-format=1.3] S[table-format=1.3] S[table-format=1.3] S[table-format=1.3] S[table-format=1.3] S[table-format=1.3] S[table-format=1.3]}
    \toprule
    \textbf{Pipeline} &
    \multicolumn{1}{c}{\textbf{MMLU}} &
    \multicolumn{1}{c}{\makecell{\textbf{ARC-}\\\textbf{Easy}}} &
    \multicolumn{1}{c}{\makecell{\textbf{ARC-}\\\textbf{Challenge}}} &
    \multicolumn{1}{c}{\textbf{CSQA}} &
    \multicolumn{1}{c}{\makecell{\textbf{Hella}\\\textbf{Swag}}} &
    \multicolumn{1}{c}{\makecell{\textbf{OpenBook}\\\textbf{QA}}} &
    \multicolumn{1}{c}{\textbf{PIQA}} &
    \multicolumn{1}{c}{\makecell{\textbf{Social}\\\textbf{IQA}}} &
    \multicolumn{1}{c}{\makecell{\textbf{Wino}\\\textbf{Grande}}} &
    \multicolumn{1}{c}{\textbf{Aggregate}} \\
    \midrule
    \multicolumn{11}{l}{\textbf{(1B scale; 20B tokens)}} \\
    \midrule
    FineWeb & 0.294729 & 0.599 & 0.313993 & 0.503686 & 0.524 & 0.386 & 0.691 & 0.486 & 0.534333 & 0.481415667 \\
    DCLM & 0.312332 & {\bfseries\num{0.686}} & 0.361775 & 0.559378 & 0.541 & 0.448 & 0.696 & 0.496 & {\bfseries\num{0.562747}} & 0.518136889 \\
    \rowcolor{blue!10} \textbf{FLUX} & {\bfseries\num{0.316814}} & 0.671 & {\bfseries\num{0.363481}} & {\bfseries\num{0.591319}} & {\bfseries\num{0.554}} & {\bfseries\num{0.496}} & {\bfseries\num{0.728}} & {\bfseries\num{0.498}} & 0.540647 & {\bfseries\num{0.5288067778}} \\
    \midrule
    \multicolumn{11}{l}{\textbf{(530M scale; 10.6B tokens)}} \\
    \midrule
    FineWeb & 0.280 & 0.545 & 0.283 & 0.453 & 0.418 & 0.360 & 0.670 & 0.447 & 0.512 & 0.4408 \\
    DCLM & 0.296 & 0.618 & {\bfseries\num{0.330}} & 0.521 & 0.449 & 0.386 & 0.662 & 0.454 & 0.527 & 0.4714 \\
    \rowcolor{blue!10} \textbf{FLUX} & {\bfseries\num{0.298}} & {\bfseries\num{0.639}} & 0.315 & {\bfseries\num{0.527}} & {\bfseries\num{0.461}} & {\bfseries\num{0.390}} & {\bfseries\num{0.697}} & {\bfseries\num{0.480}} & {\bfseries\num{0.532}} & {\bfseries\num{0.4821}} \\
    \bottomrule
  \end{tabular}
  \caption{\textbf{Results across the nine-benchmark suite.} FLUX outperforms both DCLM and FineWeb on aggregate score at both the 1B and 530M scales, confirming that the quality advantage observed at 3B is consistent across compute regimes.}
  \label{tab:1b_530m_results}
\end{table}


\begin{thebibliography}{53}
\providecommand{\natexlab}[1]{#1}
\providecommand{\url}[1]{\texttt{#1}}
\expandafter\ifx\csname urlstyle\endcsname\relax
  \providecommand{\doi}[1]{doi: #1}\else
  \providecommand{\doi}{doi: \begingroup \urlstyle{rm}\Url}\fi

\bibitem[Raffel et~al.(2023)Raffel, Shazeer, Roberts, Lee, Narang, Matena, Zhou, Li, and Liu]{raffel_exploring_2023}
Colin Raffel, Noam Shazeer, Adam Roberts, Katherine Lee, Sharan Narang, Michael Matena, Yanqi Zhou, Wei Li, and Peter~J. Liu.
\newblock Exploring the {Limits} of {Transfer} {Learning} with a {Unified} {Text}-to-{Text} {Transformer}, September 2023.
\newblock URL \url{http://arxiv.org/abs/1910.10683}.
\newblock arXiv:1910.10683 [cs].

\bibitem[Hoffmann et~al.(2022)Hoffmann, Borgeaud, Mensch, Buchatskaya, Cai, Rutherford, Casas, Hendricks, Welbl, Clark, Hennigan, Noland, Millican, Driessche, Damoc, Guy, Osindero, Simonyan, Elsen, Rae, Vinyals, and Sifre]{hoffmann_training_2022}
Jordan Hoffmann, Sebastian Borgeaud, Arthur Mensch, Elena Buchatskaya, Trevor Cai, Eliza Rutherford, Diego de~Las Casas, Lisa~Anne Hendricks, Johannes Welbl, Aidan Clark, Tom Hennigan, Eric Noland, Katie Millican, George van~den Driessche, Bogdan Damoc, Aurelia Guy, Simon Osindero, Karen Simonyan, Erich Elsen, Jack~W. Rae, Oriol Vinyals, and Laurent Sifre.
\newblock Training {Compute}-{Optimal} {Large} {Language} {Models}, March 2022.
\newblock URL \url{http://arxiv.org/abs/2203.15556}.
\newblock arXiv:2203.15556 [cs].

\bibitem[Kaplan et~al.(2020)Kaplan, McCandlish, Henighan, Brown, Chess, Child, Gray, Radford, Wu, and Amodei]{kaplan_scaling_2020}
Jared Kaplan, Sam McCandlish, Tom Henighan, Tom~B. Brown, Benjamin Chess, Rewon Child, Scott Gray, Alec Radford, Jeffrey Wu, and Dario Amodei.
\newblock Scaling {Laws} for {Neural} {Language} {Models}, January 2020.
\newblock URL \url{http://arxiv.org/abs/2001.08361}.
\newblock arXiv:2001.08361 [cs].

\bibitem[Lee et~al.(2022)Lee, Ippolito, Nystrom, Zhang, Eck, Callison-Burch, and Carlini]{lee_deduplicating_2022}
Katherine Lee, Daphne Ippolito, Andrew Nystrom, Chiyuan Zhang, Douglas Eck, Chris Callison-Burch, and Nicholas Carlini.
\newblock Deduplicating {Training} {Data} {Makes} {Language} {Models} {Better}, March 2022.
\newblock URL \url{http://arxiv.org/abs/2107.06499}.
\newblock arXiv:2107.06499 [cs].

\bibitem[Soldaini et~al.(2024)Soldaini, Kinney, Bhagia, Schwenk, Atkinson, Authur, Bogin, Chandu, Dumas, Elazar, Hofmann, Jha, Kumar, Lucy, Lyu, Lambert, Magnusson, Morrison, Muennighoff, Naik, Nam, Peters, Ravichander, Richardson, Shen, Strubell, Subramani, Tafjord, Walsh, Zettlemoyer, Smith, Hajishirzi, Beltagy, Groeneveld, Dodge, and Lo]{soldaini_dolma_2024}
Luca Soldaini, Rodney Kinney, Akshita Bhagia, Dustin Schwenk, David Atkinson, Russell Authur, Ben Bogin, Khyathi Chandu, Jennifer Dumas, Yanai Elazar, Valentin Hofmann, Ananya~Harsh Jha, Sachin Kumar, Li~Lucy, Xinxi Lyu, Nathan Lambert, Ian Magnusson, Jacob Morrison, Niklas Muennighoff, Aakanksha Naik, Crystal Nam, Matthew~E. Peters, Abhilasha Ravichander, Kyle Richardson, Zejiang Shen, Emma Strubell, Nishant Subramani, Oyvind Tafjord, Pete Walsh, Luke Zettlemoyer, Noah~A. Smith, Hannaneh Hajishirzi, Iz~Beltagy, Dirk Groeneveld, Jesse Dodge, and Kyle Lo.
\newblock Dolma: an {Open} {Corpus} of {Three} {Trillion} {Tokens} for {Language} {Model} {Pretraining} {Research}, June 2024.
\newblock URL \url{http://arxiv.org/abs/2402.00159}.
\newblock arXiv:2402.00159 [cs].

\bibitem[Penedo et~al.(2024)Penedo, Kydlíček, allal, Lozhkov, Mitchell, Raffel, Werra, and Wolf]{penedo_fineweb_2024}
Guilherme Penedo, Hynek Kydlíček, Loubna~Ben allal, Anton Lozhkov, Margaret Mitchell, Colin Raffel, Leandro~Von Werra, and Thomas Wolf.
\newblock The {FineWeb} {Datasets}: {Decanting} the {Web} for the {Finest} {Text} {Data} at {Scale}, October 2024.
\newblock URL \url{http://arxiv.org/abs/2406.17557}.
\newblock arXiv:2406.17557 [cs].

\bibitem[Li et~al.(2025)Li, Fang, Smyrnis, Ivgi, Jordan, Gadre, Bansal, Guha, Keh, Arora, Garg, Xin, Muennighoff, Heckel, Mercat, Chen, Gururangan, Wortsman, Albalak, Bitton, Nezhurina, Abbas, Hsieh, Ghosh, Gardner, Kilian, Zhang, Shao, Pratt, Sanyal, Ilharco, Daras, Marathe, Gokaslan, Zhang, Chandu, Nguyen, Vasiljevic, Kakade, Song, Sanghavi, Faghri, Oh, Zettlemoyer, Lo, El-Nouby, Pouransari, Toshev, Wang, Groeneveld, Soldaini, Koh, Jitsev, Kollar, Dimakis, Carmon, Dave, Schmidt, and Shankar]{li_datacomp-lm_2025}
Jeffrey Li, Alex Fang, Georgios Smyrnis, Maor Ivgi, Matt Jordan, Samir Gadre, Hritik Bansal, Etash Guha, Sedrick Keh, Kushal Arora, Saurabh Garg, Rui Xin, Niklas Muennighoff, Reinhard Heckel, Jean Mercat, Mayee Chen, Suchin Gururangan, Mitchell Wortsman, Alon Albalak, Yonatan Bitton, Marianna Nezhurina, Amro Abbas, Cheng-Yu Hsieh, Dhruba Ghosh, Josh Gardner, Maciej Kilian, Hanlin Zhang, Rulin Shao, Sarah Pratt, Sunny Sanyal, Gabriel Ilharco, Giannis Daras, Kalyani Marathe, Aaron Gokaslan, Jieyu Zhang, Khyathi Chandu, Thao Nguyen, Igor Vasiljevic, Sham Kakade, Shuran Song, Sujay Sanghavi, Fartash Faghri, Sewoong Oh, Luke Zettlemoyer, Kyle Lo, Alaaeldin El-Nouby, Hadi Pouransari, Alexander Toshev, Stephanie Wang, Dirk Groeneveld, Luca Soldaini, Pang~Wei Koh, Jenia Jitsev, Thomas Kollar, Alexandros~G. Dimakis, Yair Carmon, Achal Dave, Ludwig Schmidt, and Vaishaal Shankar.
\newblock {DataComp}-{LM}: {In} search of the next generation of training sets for language models, April 2025.
\newblock URL \url{http://arxiv.org/abs/2406.11794}.
\newblock arXiv:2406.11794 [cs].

\bibitem[Gowtham et~al.(2025)Gowtham, Rupesh, Kumar, Saravanan, and Chaithanya]{gowtham_blu-werp_2025}
Gowtham, Sai Rupesh, Sanjay Kumar, Saravanan, and Venkata Chaithanya.
\newblock Blu-{WERP} ({Web} {Extraction} and {Refinement} {Pipeline}): {A} {Scalable} {Pipeline} for {Preprocessing} {Large} {Language} {Model} {Datasets}, November 2025.
\newblock URL \url{http://arxiv.org/abs/2511.18054}.
\newblock arXiv:2511.18054 [cs].

\bibitem[Groeneveld et~al.(2024)Groeneveld, Beltagy, Walsh, Bhagia, Kinney, Tafjord, Jha, Ivison, Magnusson, Wang, Arora, Atkinson, Authur, Chandu, Cohan, Dumas, Elazar, Gu, Hessel, Khot, Merrill, Morrison, Muennighoff, Naik, Nam, Peters, Pyatkin, Ravichander, Schwenk, Shah, Smith, Strubell, Subramani, Wortsman, Dasigi, Lambert, Richardson, Zettlemoyer, Dodge, Lo, Soldaini, Smith, and Hajishirzi]{groeneveld_olmo_2024}
Dirk Groeneveld, Iz~Beltagy, Pete Walsh, Akshita Bhagia, Rodney Kinney, Oyvind Tafjord, Ananya~Harsh Jha, Hamish Ivison, Ian Magnusson, Yizhong Wang, Shane Arora, David Atkinson, Russell Authur, Khyathi~Raghavi Chandu, Arman Cohan, Jennifer Dumas, Yanai Elazar, Yuling Gu, Jack Hessel, Tushar Khot, William Merrill, Jacob Morrison, Niklas Muennighoff, Aakanksha Naik, Crystal Nam, Matthew~E. Peters, Valentina Pyatkin, Abhilasha Ravichander, Dustin Schwenk, Saurabh Shah, Will Smith, Emma Strubell, Nishant Subramani, Mitchell Wortsman, Pradeep Dasigi, Nathan Lambert, Kyle Richardson, Luke Zettlemoyer, Jesse Dodge, Kyle Lo, Luca Soldaini, Noah~A. Smith, and Hannaneh Hajishirzi.
\newblock {OLMo}: {Accelerating} the {Science} of {Language} {Models}, June 2024.
\newblock URL \url{http://arxiv.org/abs/2402.00838}.
\newblock arXiv:2402.00838 [cs].

\bibitem[Gohari et~al.(2025)Gohari, Kadhe, Shah, Adam, Adebayo, Adusumilli, Ahmed, Angel, Borse, Chang, Dang, Desai, Eres, Iwamoto, Karve, Koyfman, Lee, Liu, Lublinsky, Ohko, Pesce, Touma, Wang, Witherspoon, Woisetschläger, Wood, Wu, Yoshida, Zawad, Zerfos, Zhou, and Bhattacharjee]{gohari_gneissweb_2025}
Hajar~Emami Gohari, Swanand~Ravindra Kadhe, Syed~Yousaf Shah, Constantin Adam, Abdulhamid Adebayo, Praneet Adusumilli, Farhan Ahmed, Nathalie~Baracaldo Angel, Santosh~Subhashrao Borse, Yuan-Chi Chang, Xuan-Hong Dang, Nirmit Desai, Revital Eres, Ran Iwamoto, Alexei Karve, Yan Koyfman, Wei-Han Lee, Changchang Liu, Boris Lublinsky, Takuyo Ohko, Pablo Pesce, Maroun Touma, Shiqiang Wang, Shalisha Witherspoon, Herbert Woisetschläger, David Wood, Kun-Lung Wu, Issei Yoshida, Syed Zawad, Petros Zerfos, Yi~Zhou, and Bishwaranjan Bhattacharjee.
\newblock {GneissWeb}: {Preparing} {High} {Quality} {Data} for {LLMs} at {Scale}, July 2025.
\newblock URL \url{http://arxiv.org/abs/2502.14907}.
\newblock arXiv:2502.14907 [cs].

\bibitem[Barbaresi(2021)]{barbaresi_trafilatura_2021}
Adrien Barbaresi.
\newblock Trafilatura: A web scraping library and command-line tool for text discovery and extraction.
\newblock \emph{Proceedings of the ACM Web Conference 2021}, 2021.
\newblock URL \url{https://doi.org/10.1145/3442442.3452052}.

\bibitem[Bevendorff et~al.(2021)Bevendorff, Potthast, Hagen, and Stein]{bevendorff_elastic_chatnoir_2021}
Janek Bevendorff, Martin Potthast, Matthias Hagen, and Benno Stein.
\newblock Elastic chatnoir.
\newblock Web service and documentation, 2021.
\newblock URL \url{https://www.chatnoir.eu/}.

\bibitem[Joulin et~al.(2016)Joulin, Grave, Bojanowski, and Mikolov]{joulin_bag_2016}
Armand Joulin, Edouard Grave, Piotr Bojanowski, and Tomas Mikolov.
\newblock Bag of {Tricks} for {Efficient} {Text} {Classification}, August 2016.
\newblock URL \url{http://arxiv.org/abs/1607.01759}.
\newblock arXiv:1607.01759 [cs].

\bibitem[Penedo et~al.(2023)Penedo, Malartic, Hesslow, Cojocaru, Cappelli, Alobeidli, Pannier, Almazrouei, and Launay]{penedo_refinedweb_2023}
Guilherme Penedo, Quentin Malartic, Daniel Hesslow, Ruxandra Cojocaru, Alessandro Cappelli, Hamza Alobeidli, Baptiste Pannier, Ebtesam Almazrouei, and Julien Launay.
\newblock The {RefinedWeb} {Dataset} for {Falcon} {LLM}: {Outperforming} {Curated} {Corpora} with {Web} {Data}, and {Web} {Data} {Only}, June 2023.
\newblock URL \url{http://arxiv.org/abs/2306.01116}.
\newblock arXiv:2306.01116 [cs].

\bibitem[Joulin et~al.(2017)Joulin, Grave, Bojanowski, and Mikolov]{joulin_fasttextzip_2017}
Armand Joulin, Edouard Grave, Piotr Bojanowski, and Tomas Mikolov.
\newblock Bag of tricks for efficient text classification (fasttext).
\newblock arXiv preprint arXiv:1607.01759, 2017.
\newblock URL \url{https://arxiv.org/abs/1607.01759}.

\bibitem[Rae et~al.(2021)Rae, Borgeaud, Cai, Millican, Hoffmann, Song, Aslanides, Henderson, Ring, Young, et~al.]{gopher2021}
Jack~W. Rae, Sebastian Borgeaud, Trevor Cai, Katie Millican, Jordan Hoffmann, Francis Song, John Aslanides, Sarah Henderson, Roman Ring, Susannah Young, et~al.
\newblock Scaling language models: Methods, analysis and insights from training gopher.
\newblock arXiv preprint arXiv:2112.11446, 2021.
\newblock URL \url{https://arxiv.org/abs/2112.11446}.

\bibitem[noa(2026)]{noauthor_chatnoir-euchatnoir-resiliparse_2026}
chatnoir-eu/chatnoir-resiliparse, March 2026.
\newblock URL \url{https://github.com/chatnoir-eu/chatnoir-resiliparse}.
\newblock original-date: 2021-06-22T09:03:44Z.

\bibitem[noa(2025)]{noauthor_allenaibff_2025}
allenai/bff, November 2025.
\newblock URL \url{https://github.com/allenai/bff}.
\newblock original-date: 2023-03-24T20:51:52Z.

\bibitem[Teknium(2023)]{teknium_openhermes_2_5}
Teknium.
\newblock Openhermes 2.5.
\newblock Hugging Face model card, 2023.
\newblock URL \url{https://huggingface.co/teknium/OpenHermes-2.5-Mistral-7B}.

\bibitem[Fan et~al.(2019)Fan, Jernite, Perez, Grangier, Weston, and Auli]{fan_eli5_2019}
Angela Fan, Yacine Jernite, Ethan Perez, David Grangier, Jason Weston, and Michael Auli.
\newblock Eli5: Long form question answering.
\newblock In \emph{Proceedings of the 57th Annual Meeting of the Association for Computational Linguistics}, 2019.
\newblock URL \url{https://arxiv.org/abs/1907.09190}.

\bibitem[Bloom(1970)]{bloom_spacetime_1970}
Burton~H. Bloom.
\newblock Space/time trade-offs in hash coding with allowable errors.
\newblock \emph{Communications of the ACM}, 13\penalty0 (7):\penalty0 422--426, July 1970.
\newblock ISSN 0001-0782, 1557-7317.
\newblock \doi{10.1145/362686.362692}.
\newblock URL \url{https://dl.acm.org/doi/10.1145/362686.362692}.

\bibitem[Broder(1997)]{broder1997resemblance}
Andrei~Z. Broder.
\newblock On the resemblance and containment of documents.
\newblock In \emph{Compression and Complexity of Sequences}, 1997.
\newblock URL \url{https://doi.org/10.1109/SEQUEN.1997.666900}.

\bibitem[Mizrahi et~al.(2025)Mizrahi, Larsen, Allardice, Petryk, Gorokhov, Li, Fang, Gardner, Gunter, and Dehghan]{mizrahi_language_2025}
David Mizrahi, Anders Boesen~Lindbo Larsen, Jesse Allardice, Suzie Petryk, Yuri Gorokhov, Jeffrey Li, Alex Fang, Josh Gardner, Tom Gunter, and Afshin Dehghan.
\newblock Language {Models} {Improve} {When} {Pretraining} {Data} {Matches} {Target} {Tasks}, July 2025.
\newblock URL \url{http://arxiv.org/abs/2507.12466}.
\newblock arXiv:2507.12466 [cs].

\bibitem[Hendrycks et~al.(2021)Hendrycks, Burns, Basart, Zou, Mazeika, Song, and Steinhardt]{hendrycks_measuring_2021}
Dan Hendrycks, Collin Burns, Steven Basart, Andy Zou, Mantas Mazeika, Dawn Song, and Jacob Steinhardt.
\newblock Measuring {Massive} {Multitask} {Language} {Understanding}, January 2021.
\newblock URL \url{http://arxiv.org/abs/2009.03300}.
\newblock arXiv:2009.03300 [cs].

\bibitem[noa()]{noauthor_hellaswag_nodate}
{HellaSwag}: {Can} a {Machine} {Really} {Finish} {Your} {Sentence}? - {ACL} {Anthology}.
\newblock URL \url{https://aclanthology.org/P19-1472/}.

\bibitem[Clark et~al.(2018)Clark, Cowhey, Etzioni, Khot, Sabharwal, Schoenick, and Tafjord]{clark_think_2018}
Peter Clark, Isaac Cowhey, Oren Etzioni, Tushar Khot, Ashish Sabharwal, Carissa Schoenick, and Oyvind Tafjord.
\newblock Think you have {Solved} {Question} {Answering}? {Try} {ARC}, the {AI2} {Reasoning} {Challenge}, March 2018.
\newblock URL \url{http://arxiv.org/abs/1803.05457}.
\newblock arXiv:1803.05457 [cs].

\bibitem[Talmor et~al.(2019)Talmor, Herzig, Lourie, and Berant]{talmor_commonsenseqa_2019}
Alon Talmor, Jonathan Herzig, Nicholas Lourie, and Jonathan Berant.
\newblock {CommonsenseQA}: {A} {Question} {Answering} {Challenge} {Targeting} {Commonsense} {Knowledge}, March 2019.
\newblock URL \url{http://arxiv.org/abs/1811.00937}.
\newblock arXiv:1811.00937 [cs].

\bibitem[Bisk et~al.(2019)Bisk, Zellers, Bras, Gao, and Choi]{bisk_piqa_2019}
Yonatan Bisk, Rowan Zellers, Ronan~Le Bras, Jianfeng Gao, and Yejin Choi.
\newblock {PIQA}: {Reasoning} about {Physical} {Commonsense} in {Natural} {Language}, November 2019.
\newblock URL \url{http://arxiv.org/abs/1911.11641}.
\newblock arXiv:1911.11641 [cs].

\bibitem[Sap et~al.(2019)Sap, Rashkin, Chen, LeBras, and Choi]{sap_socialiqa_2019}
Maarten Sap, Hannah Rashkin, Derek Chen, Ronan LeBras, and Yejin Choi.
\newblock {SocialIQA}: {Commonsense} {Reasoning} about {Social} {Interactions}, September 2019.
\newblock URL \url{http://arxiv.org/abs/1904.09728}.
\newblock arXiv:1904.09728 [cs].

\bibitem[Sakaguchi et~al.(2019)Sakaguchi, Bras, Bhagavatula, and Choi]{sakaguchi_winogrande_2019}
Keisuke Sakaguchi, Ronan~Le Bras, Chandra Bhagavatula, and Yejin Choi.
\newblock {WinoGrande}: {An} {Adversarial} {Winograd} {Schema} {Challenge} at {Scale}, November 2019.
\newblock URL \url{http://arxiv.org/abs/1907.10641}.
\newblock arXiv:1907.10641 [cs].

\bibitem[Mihaylov et~al.(2018)Mihaylov, Clark, Khot, and Sabharwal]{mihaylov_can_2018}
Todor Mihaylov, Peter Clark, Tushar Khot, and Ashish Sabharwal.
\newblock Can a {Suit} of {Armor} {Conduct} {Electricity}? {A} {New} {Dataset} for {Open} {Book} {Question} {Answering}, September 2018.
\newblock URL \url{http://arxiv.org/abs/1809.02789}.
\newblock arXiv:1809.02789 [cs].

\bibitem[Merrick et~al.(2024)Merrick, Jain, et~al.]{merrick_arctic-embed_2024}
Luke Merrick, Abhinav Jain, et~al.
\newblock Arctic-embed: Scalable, efficient, and accurate text embedding models.
\newblock GitHub repository and technical report, 2024.
\newblock URL \url{https://github.com/Snowflake-Labs/arctic-embed}.

\bibitem[Jacovi et~al.(2023)Jacovi, Caciularu, Goldman, and Goldberg]{jacovi_stop_2023}
Alon Jacovi, Avi Caciularu, Omer Goldman, and Yoav Goldberg.
\newblock Stop {Uploading} {Test} {Data} in {Plain} {Text}: {Practical} {Strategies} for {Mitigating} {Data} {Contamination} by {Evaluation} {Benchmarks}, October 2023.
\newblock URL \url{http://arxiv.org/abs/2305.10160}.
\newblock arXiv:2305.10160 [cs].

\bibitem[{Allen Institute for AI}(2024)]{allenai2024decon}
{Allen Institute for AI}.
\newblock decon: A toolkit for dataset decontamination.
\newblock GitHub repository, 2024.
\newblock URL \url{https://github.com/allenai/decon}.

\bibitem[Zellers et~al.(2019)Zellers, Holtzman, Bisk, Farhadi, and Choi]{zellers_hellaswag_2019}
Rowan Zellers, Ari Holtzman, Yonatan Bisk, Ali Farhadi, and Yejin Choi.
\newblock {HellaSwag}: {Can} a {Machine} {Really} {Finish} {Your} {Sentence}?, May 2019.
\newblock URL \url{http://arxiv.org/abs/1905.07830}.
\newblock arXiv:1905.07830 [cs].

\bibitem[Radford et~al.()Radford, Narasimhan, Salimans, and Sutskever]{radford_improving_nodate}
Alec Radford, Karthik Narasimhan, Tim Salimans, and Ilya Sutskever.
\newblock Improving {Language} {Understanding} by {Generative} {Pre}-{Training}.

\bibitem[Touvron et~al.(2023)Touvron, Lavril, Izacard, Martinet, Lachaux, Lacroix, Rozière, Goyal, Hambro, Azhar, Rodriguez, Joulin, Grave, and Lample]{touvron_llama_2023}
Hugo Touvron, Thibaut Lavril, Gautier Izacard, Xavier Martinet, Marie-Anne Lachaux, Timothée Lacroix, Baptiste Rozière, Naman Goyal, Eric Hambro, Faisal Azhar, Aurelien Rodriguez, Armand Joulin, Edouard Grave, and Guillaume Lample.
\newblock {LLaMA}: {Open} and {Efficient} {Foundation} {Language} {Models}, February 2023.
\newblock URL \url{http://arxiv.org/abs/2302.13971}.
\newblock arXiv:2302.13971 [cs].

\bibitem[Paszke et~al.(2019)Paszke, Gross, Massa, Lerer, Bradbury, Chanan, Killeen, Lin, Gimelshein, Antiga, Desmaison, Köpf, Yang, DeVito, Raison, Tejani, Chilamkurthy, Steiner, Fang, Bai, and Chintala]{paszke_pytorch_2019}
Adam Paszke, Sam Gross, Francisco Massa, Adam Lerer, James Bradbury, Gregory Chanan, Trevor Killeen, Zeming Lin, Natalia Gimelshein, Luca Antiga, Alban Desmaison, Andreas Köpf, Edward Yang, Zach DeVito, Martin Raison, Alykhan Tejani, Sasank Chilamkurthy, Benoit Steiner, Lu~Fang, Junjie Bai, and Soumith Chintala.
\newblock {PyTorch}: {An} {Imperative} {Style}, {High}-{Performance} {Deep} {Learning} {Library}, December 2019.
\newblock URL \url{http://arxiv.org/abs/1912.01703}.
\newblock arXiv:1912.01703 [cs].

\bibitem[Li et~al.(2020)Li, Zhao, Varma, Salpekar, Noordhuis, Li, Paszke, Smith, Vaughan, Damania, and Chintala]{li_pytorch_2020}
Shen Li, Yanli Zhao, Rohan Varma, Omkar Salpekar, Pieter Noordhuis, Teng Li, Adam Paszke, Jeff Smith, Brian Vaughan, Pritam Damania, and Soumith Chintala.
\newblock {PyTorch} {Distributed}: {Experiences} on {Accelerating} {Data} {Parallel} {Training}, June 2020.
\newblock URL \url{http://arxiv.org/abs/2006.15704}.
\newblock arXiv:2006.15704 [cs].

\bibitem[Micikevicius et~al.(2018)Micikevicius, Narang, Alben, Diamos, Elsen, Garcia, Ginsburg, Houston, Kuchaiev, Venkatesh, and Wu]{micikevicius_mixed_2018}
Paulius Micikevicius, Sharan Narang, Jonah Alben, Gregory Diamos, Erich Elsen, David Garcia, Boris Ginsburg, Michael Houston, Oleksii Kuchaiev, Ganesh Venkatesh, and Hao Wu.
\newblock Mixed {Precision} {Training}, February 2018.
\newblock URL \url{http://arxiv.org/abs/1710.03740}.
\newblock arXiv:1710.03740 [cs].

\bibitem[Zhang and Sennrich(2019)]{zhang2019rmsnorm}
Biao Zhang and Rico Sennrich.
\newblock Root mean square layer normalization.
\newblock In \emph{Advances in Neural Information Processing Systems}, 2019.
\newblock URL \url{https://arxiv.org/abs/1910.07467}.

\bibitem[Shazeer(2020)]{shazeer2020glu}
Noam Shazeer.
\newblock Glu variants improve transformer.
\newblock arXiv preprint arXiv:2002.05202, 2020.
\newblock URL \url{https://arxiv.org/abs/2002.05202}.

\bibitem[Su et~al.(2023)Su, Lu, Pan, Murtadha, Wen, and Liu]{su_roformer_2023}
Jianlin Su, Yu~Lu, Shengfeng Pan, Ahmed Murtadha, Bo~Wen, and Yunfeng Liu.
\newblock {RoFormer}: {Enhanced} {Transformer} with {Rotary} {Position} {Embedding}, November 2023.
\newblock URL \url{http://arxiv.org/abs/2104.09864}.
\newblock arXiv:2104.09864 [cs].

\bibitem[Dao et~al.(2022)Dao, Fu, Ermon, Rudra, and Ré]{dao_flashattention_2022}
Tri Dao, Daniel~Y. Fu, Stefano Ermon, Atri Rudra, and Christopher Ré.
\newblock {FlashAttention}: {Fast} and {Memory}-{Efficient} {Exact} {Attention} with {IO}-{Awareness}, June 2022.
\newblock URL \url{http://arxiv.org/abs/2205.14135}.
\newblock arXiv:2205.14135 [cs].

\bibitem[Glorot and Bengio(2010)]{glorot_understanding_2010}
Xavier Glorot and Yoshua Bengio.
\newblock Understanding the difficulty of training deep feedforward neural networks.
\newblock In \emph{Proceedings of the {Thirteenth} {International} {Conference} on {Artificial} {Intelligence} and {Statistics}}, pages 249--256. JMLR Workshop and Conference Proceedings, March 2010.
\newblock URL \url{https://proceedings.mlr.press/v9/glorot10a.html}.

\bibitem[{EleutherAI}(2022)]{gptneox2022}
{EleutherAI}.
\newblock Gpt-neox-20b.
\newblock GitHub repository, 2022.
\newblock URL \url{https://github.com/EleutherAI/gpt-neox}.

\bibitem[Loshchilov and Hutter(2019)]{loshchilov2019adamw}
Ilya Loshchilov and Frank Hutter.
\newblock Decoupled weight decay regularization.
\newblock International Conference on Learning Representations (ICLR), 2019.
\newblock URL \url{https://arxiv.org/abs/1711.05101}.

\bibitem[Pascanu et~al.(2013)Pascanu, Mikolov, and Bengio]{pascanu_difficulty_2013}
Razvan Pascanu, Tomas Mikolov, and Yoshua Bengio.
\newblock On the difficulty of training {Recurrent} {Neural} {Networks}, February 2013.
\newblock URL \url{http://arxiv.org/abs/1211.5063}.
\newblock arXiv:1211.5063 [cs].

\bibitem[Loshchilov and Hutter(2017)]{loshchilov_sgdr_2017}
Ilya Loshchilov and Frank Hutter.
\newblock {SGDR}: {Stochastic} {Gradient} {Descent} with {Warm} {Restarts}, May 2017.
\newblock URL \url{http://arxiv.org/abs/1608.03983}.
\newblock arXiv:1608.03983 [cs].

\bibitem[{Hugging Face}(2023)]{lighteval2023}
{Hugging Face}.
\newblock Lighteval.
\newblock GitHub repository, 2023.
\newblock URL \url{https://github.com/huggingface/lighteval}.

\bibitem[Golubin(2026)]{selectolax_github}
Artem Golubin.
\newblock rushter/selectolax.
\newblock GitHub repository, 2026.
\newblock URL \url{https://github.com/rushter/selectolax}.

\bibitem[Prigent(2024)]{ut1_blocklist}
Fabrice Prigent.
\newblock Ut1 blacklists.
\newblock \url{https://dsi.ut-capitole.fr/blacklists/}, 2024.
\newblock Universit'e Toulouse 1 Capitole.

\bibitem[Wettig et~al.(2025)Wettig, Lo, Min, Hajishirzi, Chen, and Soldaini]{wettig_organize_2025}
Alexander Wettig, Kyle Lo, Sewon Min, Hannaneh Hajishirzi, Danqi Chen, and Luca Soldaini.
\newblock Organize the web: Constructing domains enhances pre-training data curation.
\newblock \emph{arXiv preprint arXiv:2502.10341}, 2025.

\end{thebibliography}
\end{document}